\documentclass[twoside]{article}

\usepackage[accepted]{aistats2023}
\usepackage[english]{babel}
\usepackage[square,authoryear,sort]{natbib}

\usepackage[utf8]{inputenc} 
\usepackage[T1]{fontenc}    
\usepackage{hyperref}       
\usepackage{url}            
\usepackage{booktabs}       
\usepackage{amsfonts}       
\usepackage{nicefrac}       
\usepackage{microtype}      
\usepackage{xcolor}         
\usepackage{graphicx}
\usepackage{amsmath}
\usepackage{placeins}
\usepackage{amsthm}
\usepackage{thmtools} 
\usepackage{thm-restate}
\usepackage{cleveref}
\usepackage{nameref}
\usepackage{mathtools}
\usepackage{float}
\usepackage{FiraMono}
\usepackage[perpage]{footmisc}

\definecolor{myorange}{rgb}{0.87, 0.56, 0.02}
\definecolor{myblue}{rgb}{0.00392156862745098, 0.45098039215686275, 0.980392156862745}
\definecolor{mygreen}{rgb}{0.00784313725490196, 0.6196078431372549, 0.45098039215686275}
\definecolor{myred}{rgb}{0.8352941176470589, 0.3686274509803922, 0.0}
\definecolor{mypurple2}{rgb}{0.8, 0.47058823529411764, 0.7372549019607844}
\definecolor{mypurple}{rgb}{0.501, 0, 0.501}

\newtheorem{remark}{Remark}
\newtheorem{lemma}{Lemma}

\hypersetup{
    colorlinks,
    citecolor=[rgb]{0.501, 0, 0.501},
    linkcolor=[rgb]{0.501, 0, 0.501},
    urlcolor=[rgb]{0.501, 0, 0.501},
}

\newcommand{\errbar}[1]{{\color{gray}\tiny$\pm$#1}}

\begin{document}

\twocolumn[

\aistatstitle{Do Bayesian Neural Networks Need To Be Fully Stochastic?}

\aistatsauthor{Mrinank Sharma\And Sebastian Farquhar \And Eric Nalisnick \And Tom Rainforth}
\aistatsaddress{University of Oxford \And University of Oxford \And University of Amsterdam \And University of Oxford }]

\begin{abstract}
We investigate the benefit of treating \emph{all} the parameters in a Bayesian neural network stochastically and find compelling theoretical and empirical evidence that this standard construction may be unnecessary.
To this end, we prove that expressive predictive distributions require only small amounts of stochasticity. In particular, partially stochastic networks with only $n$ stochastic biases are universal probabilistic predictors
for $n$-dimensional predictive problems.
In empirical investigations, we find no systematic benefit of full stochasticity across four different inference modalities and eight datasets; partially stochastic networks can match and sometimes even outperform fully stochastic networks, despite their reduced memory costs.
\end{abstract}

\section{Introduction}
Bayesian neural networks (BNNs) are often considered to be the most principled approach for uncertainty quantification in deep learning~\citep{mackay1992bayesian,neal2012bayesian,wilson2020case,abdar2021review}.
Indeed, they have a simple and compelling foundation: we use neural networks to define  flexible hypotheses classes of predictive functions by defining a prior over \emph{all} their weights and biases, then perform inference to produce posterior predictive distributions.

In practice, full posterior inference for BNNs is intractable and so practitioners must resort to approximate inference schemes~\citep{welling2011bayesian,neal2012bayesian, blundell2015weight, daxberger2021laplace}. 
This can lead to practical behaviour that is highly distinct from that of the true posterior \citep{foong2020expressiveness, coker2021wide}, while still being extremely computationally expensive.

\begin{figure}[t]
    \centering
    \includegraphics[width=\linewidth]{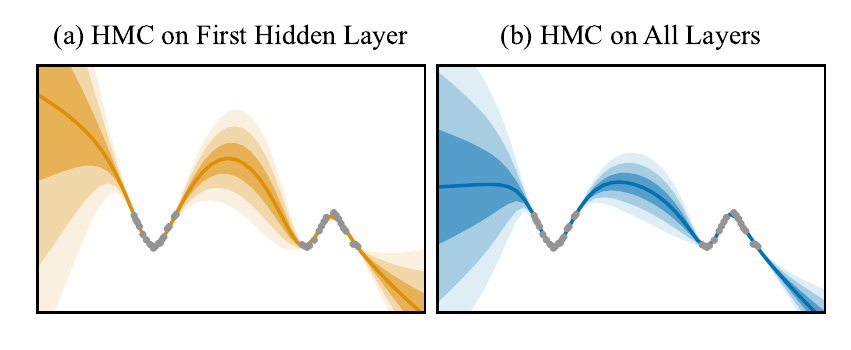}
    \vspace{-20pt}
    \caption{Perhaps surprisingly, inference over only the first hidden layer weights of a small multi-layer perceptron represents uncertainty as well as inference over all weights, whilst training c.a.~7 times faster. We first train a maximum-a-posterior network and then use Hamiltonian Monte Carlo inference over (a) the first hidden layer parameters only---other parameters are fixed---and (b) all network parameters. Lines: mean predictions. Shaded areas: predictive intervals.}
    \label{fig:toy_hmc_intro}
    \vspace{-10pt}
\end{figure}

To reduce these costs, the research community has recently considered partially stochastic networks~\citep{snoek2015scalable,ober2019benchmarking,kristiadi2020being,izmailov2020subspace,daxberger2021bayesian,daxberger2021laplace,lei2021spatial}. Though promising, these approaches are usually seen as pragmatic cost-saving measures relative to more expensive but more principled fully stochastic networks. Indeed, \citet{kristiadi2020being} describe stochastic last-layer approaches as ``approximation schemes,'' \citet{daxberger2021bayesian} see partial stochasticity as a tool for approximating the full posterior predictive, and \citet{ober2019benchmarking} describe a compromise between ``tractability and expressiveness.'' 

In this work, we question this underlying assumption that full stochasticity is preferable to, and indeed more principled than, partial stochasticity. Despite the prevalence of this assumption, we uncover compelling theoretical and empirical evidence that suggests it may be misguided. 

To begin, we first consider whether full stochasticity is necessary for our networks to be sufficiently expressive (§\ref{sec:theory}).
Although one may intuit that reducing the number of stochastic parameters hampers expressivity, we prove this is not the case.
In fact, many simple architectures using only a handful of stochastic parameters are universal conditional distribution approximators (UCDAs)---they can sample from any continuous conditional distribution arbitrarily well. Moreover, finite-width bounded-variance fully stochastic layers can even destroy information about the input. These results demonstrate full stochasticity is certainly not necessary for expressive predictive distributions.

We then question whether full stochasticity can be justified by its original Bayesian formulation by examining whether approximate inference can faithfully represent the posterior. Here, we find even state-of-the-art inference schemes using impractical amounts of compute do \textit{not} produce faithful representations (§\ref{sec:hmc_mixing}). Thus fully stochastic networks cannot be supported through their Bayesian formulation alone.

Of course, full stochasticity could still be a \textit{practically} helpful construction for learning useful predictive distributions. Accordingly, we empirically investigate whether full stochasticity translates to improved predictive performance over partially stochastic networks (§\ref{sec:experiments}). In fact, across four inference modalities and eight datasets, we find no systematic benefit of full stochasticity; partially stochastic networks can match and sometimes even outperform fully stochastic networks, despite reduced memory costs and typically shorter training times (Fig.~\ref{fig:toy_hmc_intro}). 

Overall, our work questions the prevalent assumption that full stochasticity is preferable to and more principled than partial stochasticity. We demonstrate that partially stochastic networks are no less principled than fully stochastic ones, challenging the \textit{de facto} default model construction of full stochasticity.
To summarise, our key contributions are:
\begin{enumerate}
\vspace{-5pt}
    \item[(i)] We show that there is no tradeoff between the number of stochastic parameters and network expressivity. In particular, we prove partially stochastic networks are universal conditional distribution approximators. 
    \vspace{-5pt}
    \item[(ii)] Across four inference modalities, ranging from high-fidelity Hamiltonian Monte Carlo to crude mean-field variational inference, we demonstrate that full stochasticity does not improve practical predictive performance. Surprisingly, we consistently find partially stochastic networks that match or outperform their fully stochastic variants.
    However, the best-performing partially stochastic network varies by inference modality.
\vspace{-5pt}
\end{enumerate}

\section{Background}
We focus on supervised learning problems. Let the training set be denoted as $\mathcal{D}=\{(x_i, {y}_i) \}_{i=1}^{N}$ with inputs $x_i \in \mathcal{X}$ and outputs $y_i \in \mathcal{Y}$. We assume the data is independently and identically drawn from an underlying distribution $P_{X,Y}$. Our task is to learn a conditional distribution $Y|X=x$.

\textbf{Bayesian Neural Networks (BNNs)}~ Let $f_{\theta}(x)$ be a deep neural network with parameters $\theta$, which represent a set of weights and biases.
Rather than employing empirical risk minimization to train $\theta$, BNNs place a prior $p(\theta)$ over $\theta$ and define a likelihood, $p(y|f_{\theta}(x))$.
By Bayes' rule, this now defines a \textit{posterior}, $p(\theta|{\mathcal{D}}) \propto p(\theta) p(\mathcal{D}|\theta)$---where $p(\mathcal{D}|\theta) = \prod_i p(y_i| f_{\theta}(x_i))$---that represents the updated beliefs about $\theta$ given the data $\mathcal{D}$.
Prediction is performed using the
\textit{posterior predictive}, $p(y | x, \mathcal{D}) = \mathbb{E}_{p(\theta | \mathcal{D})}\left[ p(y|f_{\theta}(x))\right]$, which represents the push forward distribution of the posterior through the network for a given input $x$.
Given that BNNs are explicitly algorithms for supervised prediction, one ultimately only cares about this posterior \textit{predictive} distribution, rather than the posterior itself \citep{foong2020expressiveness,farquhar2020liberty}.
The properties of the posterior predictive distribution are often referred to as the ``function space'' properties of a BNN~\citep{izmailov2021bayesian}.

\textbf{Approximate Inference in BNNs}~ Unfortunately, exact inference is generally intractable for BNNs. As such, practitioners resort to approximate inference, typically over all model parameters. Sampling-based approaches, such as Hamiltonian Monte Carlo (HMC) \citep{neal2012bayesian} or Stochastic Gradient Langevin Dynamics \citep{welling2011bayesian} attempt to sample from the posterior. Alternatively, traditional variational approaches \citep{mackay1992bayesian,blundell2015weight,gal2016dropout} learn an approximate posterior, $q(\theta; \phi) \approx p(\theta|\mathcal{D})$, for which existing methods usually make some kind of mean-field assumption over $\theta$.
Meanwhile, some modern approaches have instead looked directly to learn variational approximations of the posterior predictive itself~\citep{sun2019functional, ma2019variational, rudner2020rethinking, santana2021function}. 

\textbf{Partially Stochastic Networks}~ Let $f_{\Theta}(x)$ be a deep neural network and define a likelihood $p(y|f_\Theta(x))$. In a partially stochastic network \citep{kristiadi2020being,kristiadi2021learnable,daxberger2021bayesian,izmailov2020subspace,dusenberry2020efficient,lei2021spatial,snoek2015scalable}, we have $\Theta = \Theta_{S} \cup \Theta_{D}$. 
We learn point estimates for $\Theta_D$ and a distribution over $\Theta_S$, which could be learnt jointly with the deterministic parameters or separately in a two-stage training procedure.
To make predictions, we compute the \textit{subset predictive distribution} by holding $\Theta_D$ fixed and pushing forward the distribution over $\Theta_S$ through the network. 



\section{Related Work}
\textbf{Limitations of BNNs}~ Several works raise concerns with BNNs. \citet{foong2020expressiveness}, \citet{coker2021wide}, and \citet{trippe2018overpruning} showed mean-field variational inference behaves pathologically. Others find deviating from the posterior predictive---for instance, by sharpening the posterior~\citep{wenzel2020good} or degrading inference quality~\citep{izmailov2021dangers}---actually improves practical predictive performance, thereby undermining the value of the full network posterior predictive.
Our work complements these observations. Our demonstration of inaccurate inference weakens the theoretical justification for BNNs (\S\ref{sec:hmc_mixing}). Further, we find full stochasticity consistently does not improve predictive performance (\S\ref{sec:experiments}), which similarly questions the value of the full network posterior predictive.

\textbf{Existing Partially Stochastic Networks}~ Partially stochastic networks are gaining popularity. \citet{daxberger2021bayesian} approximate full network inference by performing \textit{expressive} inference over a carefully chosen subset of model weights. Further, \citet{izmailov2020subspace} perform expressive inference in an alternative probabilistic model, constructed by projecting network parameters to a low-dimensional subspace. But we demonstrate that expressive inference is not necessary in theory (\S\ref{sec:theory}) and in practice (\S\ref{sec:experiments}). Moreover, several works consider partial stochasticity as a pragmatic cost-saving measure relative to full stochasticity~\citep{snoek2015scalable,lei2021spatial,kristiadi2020being,dusenberry2020efficient}. We, however, question the value of full stochasticity and demonstrate partial stochasticity is no less justified than full stochasticity. Finally, we show stochastic output layers---the most popular approach---are typically not universal conditional distribution approximators (\S\ref{sec:theory}). 

\paragraph{Alternative Uncertainty Quantification Approaches}
Other than BNNs, there are many approaches for uncertainty quantification in deep learning~\citep{abdar2021review}. Deep ensembles are popular and peformant~\citep{lakshminarayanan2017simple}. Others use entirely deterministic methods ~\citep{van2020uncertainty,skafte2019reliable,mukhoti2021deterministic}. Further, \citet{osband2021epistemic} suggest using neural networks to approximate inference in some other probabilistic model, rather than performing inference over a neural network's weights and biases.
Our demonstration of inaccurate inference (\S\ref{sec:hmc_mixing}) supports this perspective by highlighting the challenge of accurate posterior inference. 

\section{Expressivity of Partially Stochastic Networks}\label{sec:theory}
\vspace{-5pt}
Fully stochastic networks are typically assumed to be preferable to partially stochastic networks.
We now question this assumption by examining whether fully stochastic networks are necessary for theoretical \textit{expressivity}. That is, can partially stochastic networks, in principle, approximate conditional distributions as well as fully stochastic ones? Our findings are emphatically in the affirmative: we will show that networks using only a number of random variables equal to the dimensionality of the output space are universal conditional distribution approximators.
  
Our theoretical results leverage the \textit{Noise Outsourcing Lemma} \citep{kallenberg1997foundations, austin2012exchangeable, zhou2022deep} and the Universal Approximation Theorem (UAT) \citep{leshno1993multilayer}. We start by restating these results.

\begin{lemma}[Noise Outsourcing Lemma \citep{kallenberg1997foundations, austin2012exchangeable,zhou2022deep}]\label{lemma:noise_outsourcing}
Let $X$ and $Y$ be random variables in Borel spaces $\mathcal{X}$ and $\mathcal{Y}$. For any given $m\ge1$, there exists a random variable $\eta \sim \mathcal{N}(0,I_m)$ 
and a Borel-measurable function $\tilde{f}: \mathbb{R}^m \times \mathcal{X} \rightarrow \mathcal{Y}$ such that $\eta$ is independent of $X$ and
\begin{align}
    (X, Y) = (X, \tilde{f}(\eta, X))
\end{align}
\vspace{-5pt}almost surely. Thus, $\tilde{f}(\eta, x) \sim Y|X=x,~ \forall x \in \mathcal{X}$.
\end{lemma}
The noise outsourcing lemma states that conditional distribution estimation can always be reduced to learning an appropriate function $\tilde{f}$ that maps from the input and independent noise to the output. Thus, if we can learn a $\tilde{f}$, we can sample from $Y|X=x$ simply by sampling $\eta \sim {N}(0,I_m)$ and calculating $Y=\tilde{f}(\eta,x)$. We term $\tilde{f}$ a \emph{generator function} of the conditional distribution $Y|X$ and note that it is not unique (e.g. we can always have $\eta'=-\eta$ and $\tilde{f}'(\eta',X)=\tilde{f}(-\eta',X)$).

\begin{lemma}[Universal Approximation Theorem for Arbitrary Width Networks~\citep{leshno1993multilayer}]\label{theorem:uat}
Let $\mathcal{X}$ be some compact subspace of $\mathbb{R}^d$ and let $\mathcal{Y}\subseteq \mathbb{R}^n$. 
Further, let $f_{\theta} : \mathcal{X} \to \mathcal{Y}$ be a fully connected neural network with one hidden layer of arbitrary width and a non-polynomial activation function, where $\theta \in \Theta$ represents the parameters of the network.
Then for any arbitrary continuous function $g: \mathcal{X} \rightarrow \mathcal{Y}$ and all $\varepsilon>0$,
\begin{align}
  \exists \theta\in\Theta ~:~  \sup_{x\in\mathcal{X}} \|f_\theta(x) - g(x)\| < \varepsilon,
\end{align}
provided that the network is sufficiently wide.
\end{lemma}

Informally, Lemma~\ref{theorem:uat} states that we can approximate any continuous function arbitrarily well with a sufficiently wide network, even if that network only has a single hidden layer.

We now combine these two ideas to present our main result below in Theorem~\ref{theorem:one_bias_is_all_you_need}, which shows that arbitrary-sized networks with a small fixed amount of stochasticity \emph{before} their last layer are universal conditional distribution approximators.
Specifically, we show that the following architectures with deterministic weights can approximate any continuous conditional distribution $Y|X=x$ arbitrarily well for all $x\in \mathcal{X} \subset \mathbb{R}^d$, where $Y \in \mathcal{Y}\subseteq \mathbb{R}^n$, using only a finite set of Gaussian random variables, $Z=\{Z_1,\dots,Z_m\}$, $m\ge n$, that are independent of the input $X$ and have finite mean and variance:
\begin{itemize}
    \vspace{-8pt}
    \item[{(i)}] A deterministic multi-layer perceptron (MLP) with a single hidden layer of arbitrary width; non-polynomial activation function; and which takes $[Z;X]$ as its input.
    \item[{(ii)}] An MLP with $L=2$ layers; continuous, invertible, and non-polynomial activation functions; $d$ units with deterministic biases and $m$ units with Gaussian biases in the first layer; and a second layer of arbitrary width.
    \item[{(iii)}] An MLP with $L=2$ layers; RELU activations; $2d$ units with deterministic biases and $m$ units with Gaussian random biases in the first layer; and a second layer of arbitrary width.
    \item[{(iv)}] An MLP with $L\ge 2$ layers; continuous and non-polynomial activation functions that are either invertible or RELUs; at least $2\max(d+m,n)$ units with deterministic biases in each hidden layer; finite weights and biases throughout; one non-final hidden layer with $m$ additional units with Gaussian random biases (other layers may also have additional units with random biases, alongside their $2\max(d+m,n)$ deterministic ones), and; an arbitrary number of hidden units in one of the subsequent hidden layers.
    \vspace{-8pt}
\end{itemize}
We note that the above set of architectures is by no means exhaustive, as discussed later, but is chosen to be demonstrative of how simple architectures with universal approximation properties can be.

\begin{restatable}[Universal Conditional Distribution with Finite Stochasticity]{theorem}{ucda} \label{theorem:one_bias_is_all_you_need}
Let $X$ be a random variable taking values in $\mathcal{X}$, where $\mathcal{X}$ is a compact subspace of $\mathbb{R}^d$, and let $Y$ be a random variable taking values in $\mathcal{Y}$, where $\mathcal{Y}\subseteq \mathbb{R}^n$.
Further, let $f_{\theta} : \mathbb{R}^m \times \mathcal{X} \rightarrow \mathcal{Y}$ represent one of the neural network architectures defined in (i-iv) with deterministic parameters $\theta \in \Theta$, such that, for input $X=x$, the network produces outputs $f_{\theta}(Z,x)$, where $Z=\{Z_1,\dots,Z_m\}, Z_i\in \mathbb{R}$, are the random variables in the network, which are Gaussian, independent of $X$, and have finite mean and variance.

If there exists a continuous generator function, $\tilde{f} : \mathbb{R}^m \times \mathcal{X} \rightarrow \mathcal{Y}$, for the conditional distribution $Y|X$,
then $f_{\theta}$ can approximate $Y|X$ arbitrarily well.
Formally, $\forall \varepsilon>0, \lambda < \infty$,
\begin{align}
\exists &\theta \in \Theta, V \in \mathbb{R}^{m\times m}, u \in \mathbb{R}^m: \nonumber\\ &\sup_{x\in \mathcal{X},\eta\in \mathbb{R}^m, \|\eta\| \le \lambda} \|f_{\theta}(V 
 \eta + u,x) - \tilde{f}(\eta,x)\| < \varepsilon.
\end{align}
\end{restatable}
The proof is provided in the Supplement.  At a high level, Theorem~\ref{theorem:one_bias_is_all_you_need} shows that the collection of simple partially stochastic architectures (i-iv) are \textit{Universal Conditional Distribution Approximators} (UCDAs). That is, they can form samplers which match \emph{any continuous} target conditional distribution, $Y|X=x$, arbitrarily well: in principle, they can learn to do any probabilistic predictive task perfectly.

The high-level basis for the proof is to show a) that if our network can represent $[Z;x]$ exactly in one of its hidden layers and the downstream network is a universal deterministic approximator (as per Lemma~\ref{theorem:uat}), then it forms a UCDA, and then b) that each of the architectures (i-iv) satisfy these conditions.
Note that the distribution over the random biases in these networks does not need to be learned: we only require the presence of some random noise that can be detached from the input, and the remainder of the network to be able to approximate the conditional generating function $\tilde{f}$.

Many other partially stochastic networks will also satisfy these conditions and thus form UCDAs, though it is difficult to exactly characterize this set.
In practice, we expect \emph{most} partially stochastic networks to form UCDAs, provided that they are sufficiently large, maintain some deterministic (or arbitrarily low variance) units in each layer, and have some stochasticity \emph{before} the final layer.
One could extend our results to more complex architectures, such as those that are not fully connected (e.g.~CNNs~\citep{lecun1995convolutional}) and/or which make use of skip connections  (e.g.~ResNets~\citep{he2016deep} and DenseNets~\citep{iandola2014densenet}).
One could also consider networks with arbitrary depth, rather than arbitrary width, by using other variants of the UAT \citep{lu2017expressive, kidger2020universal}.
Meanwhile, $Z$ being non-Gaussian should also be perfectly viable, provided it is measurable with respect to a $m$-dimensional Lebesgue measure with a continuous density function.

The following property is important to note in this generalization to other architectures.
\begin{remark}
\label{rem:higher_dim}
If a continuous generator function exists for independent random noise of dimension $p$, then one also exists for any higher noise dimension $q>p$.
\end{remark}
This follows directly from the fact that the generator can simply ignore some of the noise variables.
As such, we can always add more units with stochastic biases and weights to a network without undermining universality.
However, this does not necessarily mean we can \emph{replace} the existing deterministic units with stochastic ones and maintain universality.
Our results thus explicitly \emph{do not} ratify the standard BNN case, where all the weights and biases are stochastic with \emph{bounded} means and variances: our construction relies on being able to perfectly reconstruct $X$, which is typically not possible when using a fully stochastic layer.
In other words, finite-width fully stochastic layers can, in principle, destroy required information about the input. 

\textbf{Discussion of Assumptions}~ Other than considerations about the architecture itself, the key assumption made by Theorem~\ref{theorem:one_bias_is_all_you_need} is that
a \emph{continuous} generator function exists for the conditional distribution we are approximating, $Y|X$.
Thankfully, this is generally a weak assumption, analogous to the UAT's need for a continuous target.
One can think of it as a formalization of the need for the distribution $Y|X$ itself to be continuous.

Though not an explicit condition of the theorem itself, the architectures we consider further assume that the number of stochastic variables in the network $m$ is greater than or equal to the output dimension $n$.
This is because it is difficult, albeit not necessarily impossible, for a generator function to be continuous when mapping from lower-dimensional noise to a higher-dimensional output.
However, if $Y$ is measurable with respect to an $n$-dimensional Lebesgue measure, then a continuous generator function will usually exist for exactly $m=n$ dimensional noise (and thus all $m\ge n$ by Remark~\ref{rem:higher_dim}), if one exists at all.
For example, we can consider sampling each dimension of $Y$ autoregressively using the inverse cumulative density functions of the conditionals $Y_j | X, Y_{<j}$, whenever these all exist and are continuous.

\textbf{Comparison to Previous Results}~ Our results share some similarities to previous expressivity results on \emph{fully} stochastic BNNs, most notably those of~\citet{farquhar2020liberty} and \citet{foong2020expressiveness}, who argued that deep, fully stochastic, mean-field BNNs are expressive. 
Their results rely on taking some weights in the network to the zero variance limit, which means the network is no longer fully stochastic.
Thus, though their motivations, formulations, and conclusions are quite different to our own, their results are highly compatible with ours and can be viewed as indirectly hinting at the potential benefits of partially stochastic networks.

\textbf{Classification Problems}~ Classification problems have discrete $\mathcal{Y}$ that will clearly not satisfy our assumption of a continuous generator function from $\mathbb{R}^m \times \mathcal{X}$.
Thankfully, UCDA can be achieved even more easily here by simply regressing the class probabilities $P(Y=k|X=x)$ with a deterministic network, followed by making a simple draw of the class from this categorical distribution (which can be achieved with a single, one-dimensional, random draw).

\textbf{Stochastic Last-Layer Networks are \textit{not} UCDAs}~
As an aside, we also consider the expressivity of stochastic last-layer networks (a.k.a. \textit{neural linear models}). 
Such approaches are used quite commonly in practice with notable success~\citep{daxberger2021laplace, kristiadi2020being, ober2019benchmarking, snoek2015scalable}, partially because they often allow tractable inference.
However, such architectures will generally \emph{not} be UCDAs (except for classification problems) because their distributional form of $Y|X=x$ is limited to a linear mapping of the weights and biases in the last layer.
For example, if their distribution on weights and biases is Gaussian, this will induce a Gaussian distribution on $Y|X=x$ as well.
Though this certainly does not undermine the usefulness of such approaches, it does highlight that care is required in their deployment.

\FloatBarrier
\section{Does Bayesian Reasoning Support Fully Stochastic Networks?}

\begin{figure} 
    \centering
    \includegraphics[width=\linewidth]{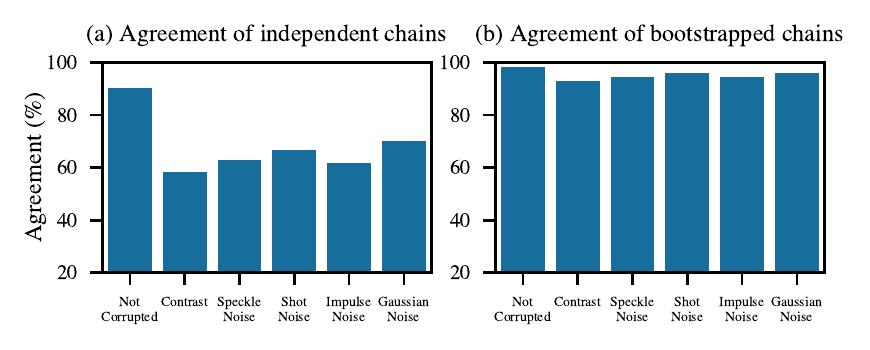}
    \vspace{-15pt}
    \caption{\textbf{Assessment of function space mixing of ResNet-20-FRN Hamiltonian Monte Carlo (HMC) samples trained on CIFAR-10.}
    We measure the variability in predictions across HMC chains released by \citet{izmailov2021bayesian}. We consider the CIFAR-10 test set and selected corruptions from the CIFAR-10-C dataset \citep{hendrycks2018benchmarking}. 
    (a) We compute the percentage of points that all three original chains make the same prediction on. 
    (b) To account for the finite sample size, we measure the variability across simulated chains formed by resampling the first HMC chain (bootstrapping). 
    The agreement of bootstrapped HMC chains is greater than 94\% across all data considered. 
    }
    \vspace{-15pt}
    \label{fig:hmc_agreement}
\end{figure}

\label{sec:hmc_mixing}
Although fully stochastic networks are unnecessary for expressive predictive distributions in theory (\S\ref{sec:theory}), full stochasticity could be supported through conformance to Bayesian principles. 
Indeed, following a strict Bayesian approach, one assumes that the observed data was generated using our probabilistic model with a fixed but unknown set of weights. Given an observed dataset, one would then place a prior distribution over all unknown parameters and perform posterior inference over each of them, which corresponds to a fully stochastic network. 
We now examine whether the purported benefits of Bayesian learning actually support the use of fully stochastic neural networks in practice.

Briefly, this strict Bayesian approach is typically justified through one or more of the following benefits: (a) the ability to naturally include prior beliefs through subjective prior distributions~\citep{neal2012bayesian}; (b) improved uncertainty estimates by averaging over different hypotheses consistent with observed data~\citep{wilson2020case}; and (c) coherent updates to uncertainty when observing data~\citep{jaynes2003probability}.

\vspace{1cm}
First, with regard to (a), standard practice is to use vague parameter-space priors~\citep{fortuin2021bayesian}. But these priors are chosen for convenience, not because they well capture our prior beliefs about the data generating process. Indeed, several studies raise serious concerns about the suitability of current BNN prior distributions~\citep{wenzel2020good,noci2021disentangling}.

Similarly, (b) does not provide support for full stochasticity. Although averaging over hypotheses consistent with observed data may improve uncertainty estimates, we do not need to use fully stochastic networks to do this. That is, we can consider different hypotheses that are consistent with observed data using partially stochastic networks.

Finally, though (c) could still support full stochasticity, it is highly dependent on our ability to perform inference accurately.
In particular, our approximations cannot be said to capture uncertainty in a ``principled'' Bayesian way if they vary significantly from true posterior.
As such, it is natural to wonder: just how challenging is accurate inference in fully stochastic networks? Can we faithfully represent the posterior distribution?

To provide some insight, we revisit the posterior samples released by \citet{izmailov2021bayesian}, who used full-batch HMC and 512 Tensor processing units---a deliberately extreme computing effort. As they do, we assess the variability of predictions across HMC chains. If each chain is well exploring the posterior predictive, the predictions made by each chain ought to agree. To assess the variability of predictions associated with the finite sample size, we resample the first HMC chain with replacement. Unlike \citet{izmailov2021bayesian}, 
we focus on out-of-distribution (OOD) data, where poor function space mixing may manifest more strongly. 

We compute the percentage of data points on which \textit{all} chains produce the same prediction.\footnote{This is different to the agreement metric of \citet{izmailov2021bayesian}, who report the percentage of data points on which one chain and the ensemble of the other two chains agree.}  As shown in Fig.~\ref{fig:hmc_agreement}{\color{mypurple}a}, while the chains agree on 90\% of the CIFAR-10 test set, the agreement falls to less than 60\% on certain OOD corruptions.
However, the agreement of the bootstrapped samples is consistently above 94\% (Fig.~\ref{fig:hmc_agreement}{\color{mypurple}b}).
The variability of predictions between chains far exceeds the variability of predictions within each chain, suggesting that each HMC chain is not well exploring the full posterior predictive distribution. Thus, additional chains would likely sample from previously unexplored regions of the posterior predictive, suggesting that the original HMC chains do not faithfully represent the posterior predictive distribution.

Even with astronomical compute and a state-of-the-art unbiased inference scheme, we see that accurate posterior inference remains elusive. 
But practical methods tend to use biased and crude posterior approximations, aggravating these concerns and leading to pathological behaviour~\citep{foong2020expressiveness,coker2021wide,trippe2018overpruning,wenzel2020good,farquhar2019unifying}.

Overall, we conclude that the use of fully stochastic methods can \textit{not} be justified by their Bayesian formulation, at least not with current inference methods.
Of course, this does not undermine the use of fully stochastic networks in and of itself. But, it does suggest adopting a holistic viewpoint, such as that of \citet{osband2021epistemic}, and focusing on developing methods that yield networks with the desired practical behaviours, rather than implicitly assuming that full approximate inference should be our ultimate aim.

\FloatBarrier
\section{Does Full Stochasticity Improve Predictions In Practice?} \label{sec:experiments}

\begin{figure}[t]
    \centering
    \includegraphics[width=\linewidth]{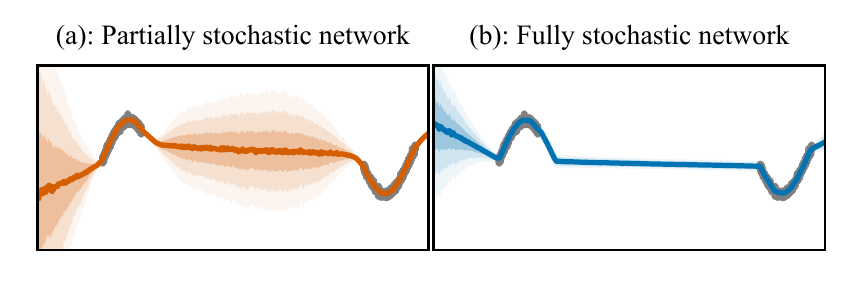}
    \vspace{-20pt}
    \caption{\textbf{1D regression with fully and partially stochastic mean-field variational inference.} The partially stochastic network has only a stochastic output layer. Lines: mean predictions. Shaded areas: $\pm\sigma, \pm2\sigma, \pm3\sigma$ predictive intervals.}
    \label{fig:vi_surprises}
    \vspace{-10pt}
\end{figure}

\begin{figure*}[t]
    \centering
    \vspace{-10pt}
    \includegraphics[width=\textwidth]{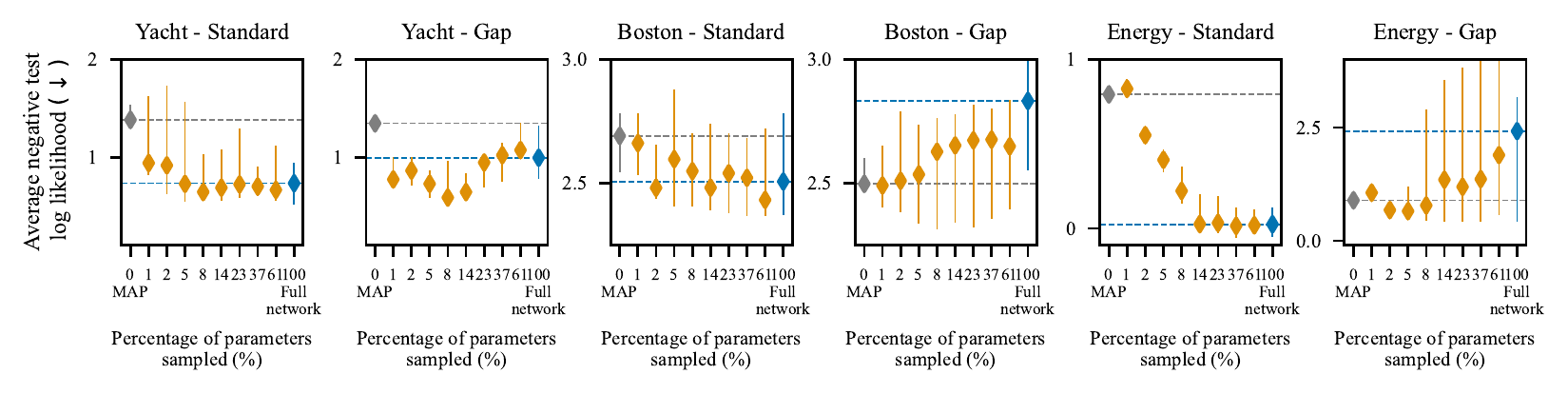}
    \vspace{-20pt}
    \caption{\textbf{UCI regression with Hamiltonian Monte Carlo (HMC)}. We use a small MLP with high-fidelity HMC inference. The partially stochastic networks first train a deterministic MAP solution, and then sample only the weights that had the largest absolute value under that MAP solution; the remaining weights are fixed at their MAP value. We consider both standard splits and gap splits \citep{foong2019between}. Diamonds: median across 15 train-test splits. Lines: interquartile range.
    }
    \label{fig:subset_hmc}
    \vspace{-5pt}
\end{figure*}

We saw that full stochasticity is unnecessary for theoretical expressivity (\S\ref{sec:theory}). Further, such networks cannot be supported through their Bayesian formulation alone (\S\ref{sec:hmc_mixing}). Nevertheless, one could hypothesize that full stochasticity is \textit{practically} useful for learning performant predictive distributions. We now examine this hypothesis: does full stochasticity improve predictive performance in practice?

Across four inference modalities and eight datasets, we find \textbf{\color{myblue}no systematic benefit of full stochasticity}. In fact, there usually exist \textbf{\color{myorange}partially stochastic networks that outperform fully stochastic ones}. 
Moreover, while previous work often argues that reducing stochasticity improves performance by enabling higher-fidelity inference~\citep{daxberger2021bayesian,izmailov2020subspace}, we show partially stochastic networks can outperform full stochastic networks, even when both networks use the same posterior approximation families over their stochastic parameters. That is, partially stochastic networks need not more expressive approximate posterior families to compensate for reduced numbers of stochastic parameters.

\textbf{Partially Stochastic Network Strategies}~ Although there are many ways to train partially stochastic networks, here, we focus on the following relatively simple strategies:
\begin{enumerate}
    \vspace{-8pt}
    \item[(i)]\textit{Two-stage training}. All parameters of the network are trained deterministically e.g., using MAP inference with prior $p_1(\Theta)=p_1(\Theta_S, \Theta_D)$. We perform (approximate) inference over the stochastic subset, targeting $p(\Theta_S|\mathcal{D}; \Theta_D)\propto p_2(\Theta_S)\prod_{i}p(y_i|f_{\Theta_S \cup \Theta_D}(x_i))$. The stochastic subset could be chosen before or after deterministic training. We could also modify the prior over $\Theta_S$ i.e., have $p_2(\Theta_S)\neq \int p_1(\Theta_S, \Theta_D)\ d\Theta_D$. Here, we consider two-stage partially stochastic variants of Hamiltonian Monte Carlo \citep{neal2012bayesian} (\S\ref{sec:1d_regression},\ref{sec:uci_regression}), Laplace Approximation \citep{mackay1992bayesian} (\S\ref{sec:vision_laplace}) and SWAG \citep{maddox2019simple} (\S\ref{sec:vision_swag}). 
    \item[(ii)]\textit{Joint training}. Alternatively, we can choose the stochastic subset \textit{a priori}, and jointly train $\Theta_D$ and $q_\Phi(\Theta_S)$. Here, we use partially stochastic variational inference  \citep{hinton1993keeping, graves2011practical,blundell2015weight} (\S\ref{sec:1d_regression},\ref{sec:vision_vi}), where $\Theta_D$ and $\Phi$ are learnt by maximising the evidence lower bound.
    \vspace{-8pt}
\end{enumerate}
We emphasise that these strategies do not directly target the full network predictive. As such, these partially stochastic networks \textit{do not} approximate the full network predictive distribution. In this section, we will examine whether their predictive distributions are useful in their own right.

\subsection{1D Regression with Hamiltonian Monte Carlo and Variational Inference}\label{sec:1d_regression}
To visually understand the effects of full and partial stochasticity, we first consider 1D regression. We consider both high-fidelity inference with Hamiltonian Monte Carlo (HMC) on a small dataset (c.a.~50 datapoints) and relatively crude approximate inference with mean-field variational inference (MFVI) on a larger dataset (c.a.~1000 datapoints). We use a two hidden layer MLP with independent $\mathcal{N}(0, \sigma^2)$ priors over the network's weights and biases.

First, on the smaller dataset, we train a deterministic MAP network. We then perform HMC over the first hidden layer weights (others fixed), and also over all weights. We follow \citet{daxberger2021bayesian} and increase the partially stochastic network's prior variance when performing HMC, also using $\sigma_\text{PS}^2 = \sigma_\text{FS}^2 \cdot |\Theta|/|\Theta_S|$.  $\sigma_\text{PS}^2$ and $\sigma_\text{FS}^2$ represent the prior variance for the partially and fully stochastic network. 

Examining the predictions (Fig.~\ref{fig:toy_hmc_intro}), we find that \textbf{\color{myblue} both networks well capture in-between uncertainty}, but the partially stochastic network trains c.a.~7 times faster. Full stochasticity does not necessarily lead to substantially improved predictions, even under high-fidelity inference.  

Second, on the larger dataset, we use MFVI to train a fully stochastic network and a partially stochastic network that uses only a stochastic output layer. We find that the fully stochastic network does not well capture in-between uncertainty (Fig.~\ref{fig:vi_surprises}{\color{mypurple}b}), even though the network is expressive enough to do so \citep{farquhar2020liberty, foong2020expressiveness}. In contrast, \textbf{\color{myorange} the partially stochastic network represents far more in-between uncertainty than the fully stochastic network} (Fig.~\ref{fig:vi_surprises}{\color{mypurple}a}), whilst also using 200 times fewer stochastic parameters. Further, both networks use \textit{the same} crude mean-field approximate posterior, showing that higher fidelity inference is not necessary for partially stochastic networks to improve performance.

\subsection{UCI Regression with Hamiltonian Monte Carlo}
\label{sec:uci_regression}
We next investigate the effect of increasing stochasticity under high-fidelity inference. That is, how does changing the number of stochastic parameters affect predictive performance? We thus use a small MLP and HMC inference on UCI regression datasets. Here, we consider partially stochastic networks with increasing numbers of stochastic parameters that are trained with two-stage HMC. That is, we first train a MAP network, and then form different stochastic networks by performing HMC over different subsets of parameters. We choose the stochastic subset by picking the weights and biases that had the maximum absolute value under the trained MAP solution. To understand the generalisation properties of these networks, we additionally consider the ``gap'' data splits from \citet{foong2019between}. To create these splits, we order the data by a chosen input feature, and use the central $10\%$ as the test set, and thus the test set represents out-of-distribution data. In contrast, the standard splits are created by uniformly sampling the dataset. For predictions, we use 600 Monte Carlo samples across 8 independent HMC chains.

We first consider how increasing stochasticity affects predictive performance on the standard splits (Fig.~\ref{fig:subset_hmc}). On these splits, we find that increasing the number of sampled parameters first improves performance, but then \textbf{\color{myblue}the benefits of further increasing stochasticity plateau}.

Furthermore, on the gap datasets, we find that \textbf{\color{myorange}increasing stochasticity first improves and then degrades performance}. This underwhelming performance of high-fidelty inference with fully stochastic BNNs on out-of-distribution (OOD) data matches observations by \citet{izmailov2021dangers}, who also found that MAP inference outperforms high-fidelty HMC on OOD data. 


Together, these results demonstrate that partially stochastic networks can match and even outperform fully stochastic networks, \textit{even when we can perform high-fidelity inference}.


\subsection{Image Classification with Laplace Approximation}
\begin{figure}
    \centering
    \includegraphics[width=\linewidth]{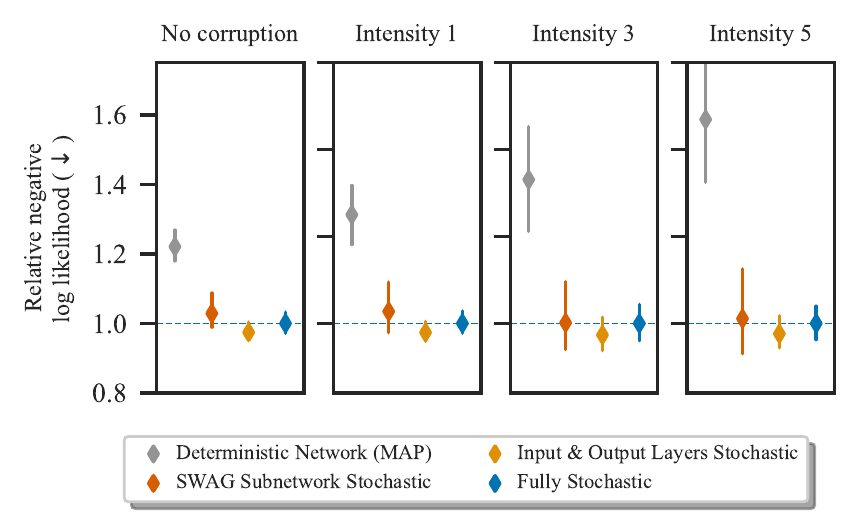}
    \vspace{-20pt}
    \caption{\textbf{Image classification with the Laplace Approximation}. We compute the average negative log-likelihood on CIFAR-10 and CIFAR-10-C relative to the fully stochastic network. Results are averaged across corruptions and shown for different corruption intensities. Markers and lines show mean and std. over 10 seeds.}
    \label{fig:laplace_results}
\end{figure}
\label{sec:vision_laplace}
We now evaluate full and partial stochasticity in larger models. To do so, we consider Laplace Approximation networks on CIFAR-10 using a WideResNet-16-4. We use two-stage training, first training a MAP solution and then using post hoc Laplace approximations on subsets of model parameters. We primarily use KFAC covariance approximations \citep{ritter2018scalable}. We also consider using a full covariance approximation using the stochastic subset selection strategy proposed by \citet{daxberger2021bayesian}---selecting parameters with the largest posterior variance under a diagonal SWAG approximation. To evaluate the networks, we compute the holdout likelihood for various networks on the CIFAR-10 and CIFAR-10-C corrupted datasets. We approximate the predictive distribution using the linearised predictive distribution \citep{immer2021improving} and the (deterministic) extended probit approximation \citep{gibbs1998bayesian}, which are the default choices suggested by \citet{daxberger2021laplace}.

When comparing the relative performance between the fully stochastic network and a partially stochastic network where only the input and output layer is stochastic (Fig.~\ref{fig:laplace_results}), we find that \textbf{\color{myorange} the partially stochastic network slightly outperforms the fully stochastic network}.\footnote{The difference in performance is statistically significant at the 5\% confidence level under a Wilcoxson signed-rank test.} 
This may be surprising since \textit{both networks use the same KFAC posterior approximation} over their stochastic parameters, but the partially stochastic network has 900 times fewer of them and predicts faster.\footnote{Although the partially stochastic network has a stochastic input layer, it is much faster than the fully stochastic network at prediction time because we use \textit{linearised} predictive distributions.}

Moreover, despite the additional costs of subnetwork selection, the increased expressivity of the posterior approximation family, and increased numbers of stochastic parameters, the `SWAG subnetwork stochastic' network actually \textit{underperforms} the stochastic input and output layer network.

\subsection{Image Classification with SWAG}
\label{sec:vision_swag}
We now investigate the effects of full and partial stochasticity under a different inference modality. We use SWA-Gaussian (SWAG, \cite{maddox2019simple}), which runs high learning rate stochastic gradient descent (SGD) starting from a set of pre-trained weights. The approximate posterior is formed by fitting a low-rank Gaussian to the SGD iterates. For the partially stochastic networks, we perform SGD only on the stochastic subset i.e., particular subsets of model parameters. We use the default hyperparameters from \citet{maddox2019simple} for SWAG with pre-trained weights, except that we tune the learning rate for each network separately. As before, we use a WideResNet-16-4 and evaluate the holdout likelihood on CIFAR-10 and CIFAR-10-C. We use 30 Monte Carlo samples when making predictions. 

\begin{figure}
    \centering
    \includegraphics[width=\linewidth]{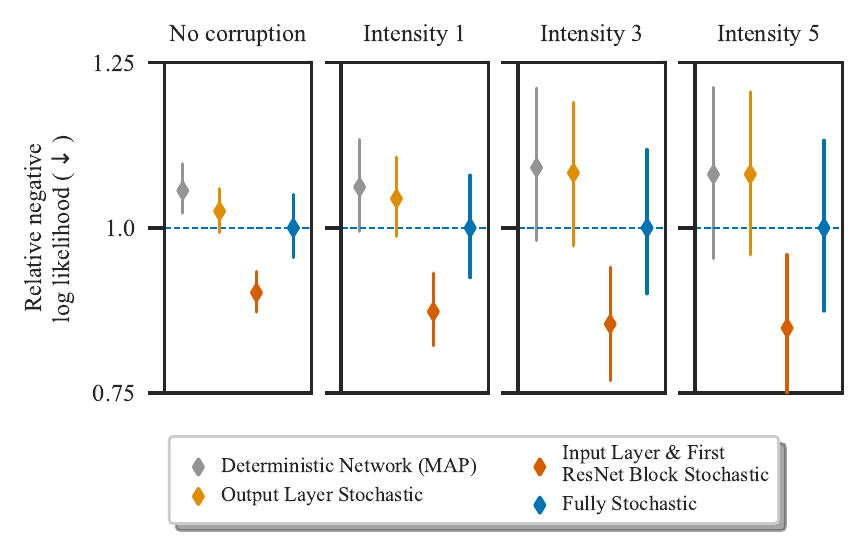}
    \vspace{-20pt}
    \caption{\textbf{Image classification with SWAG inference.} 
    We compute the average negative log-likelihood on CIFAR-10 and CIFAR-10-C relative to the fully stochastic network. Results are additionally averaged across corruptions, and shown for different corruption intensities. Markers and lines show mean and std. over 10 seeds.}
    \label{fig:swag_results}
\end{figure}

When comparing the relative performance across networks (Fig.~\ref{fig:swag_results}), we find that the fully stochastic network outperforms the deterministic network, particularly on large corruption intensities. We further find \textbf{\color{myorange}SWAG inference only over the input layer and the first ResNet block consistently outperforms the fully stochastic network}. Even though the fully stochastic network marginalises over more parameters, and thus over presumably more diverse functions, it surprisingly seems to perform worse than the partially stochastic network, despite 11x higher memory costs.

\subsection{Image Classification with Variational Inference.}
\label{sec:vision_vi}
Finally, we investigate the effects of full and partial stochasticity on even larger networks. We apply MFVI on CIFAR-10 and CIFAR-100 with a Wide-ResNet-28-10,
using the reference implementation from~\citet{nado2021uncertainty}. We report the accuracy and negative log-likelihood.
Strengthening our comparison, note that we re-used the tuned hyper-parameters for the fully stochastic and deterministic networks from \citet{nado2021uncertainty}, but did not tune the hyper-parameters for the partially stochastic networks. For predictions, we used 5 Monte Carlo samples.

We find the fully stochastic network performed worse than the deterministic network, despite using twice as many parameters. In contrast, even without tuned hyperparameters, \textbf{\color{myorange} the partially stochastic networks outperform the fully stochastic network}. The stochastic input layer performs best in terms of accuracy, and the network where the last block and output layer performs best in terms of NLL.
In particular, we emphasise the potential of stochastic input layers rather than the more commonly considered stochastic output layers.
In each case, the partially stochastic networks use only slightly more parameters than deterministic networks.

\begin{table}[]
\caption{\textbf{Partially and fully stochastic networks trained with mean-field variational inference.} We report the accuracy and average negative log-likelihood (NLL) on the CIFAR test set when performing subset VI and learning the remaining parameters by maximising the (penalised) ELBO. Mean and standard error shown across 3 seeds.}
\vspace{10pt}
\resizebox{\linewidth}{!}{\begin{tabular}{@{}lcccccc@{}}
\toprule
 & \multicolumn{2}{c}{CIFAR10} & \multicolumn{2}{c}{CIFAR100} \\ 
Model & Acc (\%) & NLL & Acc (\%) & NLL\\ \midrule
 Deterministic & 95.61 \errbar{0.01} & 0.187 \errbar{0.001}  & 79.33 \errbar{0.45} & 0.862 \errbar{0.014} \\
{\color{myblue}Fully stochastic} & 94.69 \errbar{0.07} & 0.214 \errbar{0.002} & 77.68 \errbar{0.29} & 0.944 \errbar{0.002}  \\
{\color{myorange}Input layer stochastic} & \textbf{95.70} \errbar{0.08} & 0.187 \errbar{0.002} & \textbf{79.49} \errbar{0.15} & 0.861 \errbar{0.021} \\
{\color{myorange} Output layer stochastic} & 95.60 \errbar{0.05} & 0.189 \errbar{0.001}& 78.92 \errbar{0.34} & 0.933 \errbar{0.010} \\
\begin{tabular}[l]{@{}l@{}}{\color{myorange}Output layer and}\\{\color{myorange}last block stochastic}\end{tabular} & 95.59 \errbar{0.08} & \textbf{0.168} \errbar{0.0005} & 79.00 \errbar{0.091} & \textbf{0.834} \errbar{0.0007} \\ \bottomrule
\end{tabular}}
\label{tab:vi_results_table}
\end{table}

\section{Discussion}
\label{sec:discussion}
We questioned the prevalent assumption that full stochasticity is preferable to and more principled than partial stochasticity. We found full stochasticity is not needed for theoretical expressivity~(\S\ref{sec:theory}). Further, across four inference modalities, we did not find full stochasticity to yield consistent improvements in predictive performance~(\S\ref{sec:experiments}). In fact, there usually existed partially stochastic networks that outperformed their corresponding fully stochastic variants.
Altogether, our results call into question full stochasticity as the \textit{de facto} default model construction. We believe partially stochastic networks are a highly promising model class that are just as principled as fully stochastic networks.
Indeed, we are excited to see future work that explores practical training pipelines for partially stochastic networks.
Furthermore, our observations around inaccurate inference in large BNNs (\S\ref{sec:hmc_mixing}) support holistic viewpoints such as those of \citet{osband2021epistemic}, which set aside posterior inference of neural network parameters, and instead focus on learning useful predictive distributions.
 
\subsubsection*{Acknowledgements}
M. Sharma was supported by the EPSRC Centre for Doctoral Training in Autonomous Intelligent Machines and Systems (EP/S024050/1). We thank Jan Brauner, Sören Mindermann, Freddie Bickford-Smith, Yee Whye Teh, and Rob Cornish for helpful feedback and discussions. We further thank the anonymous reviewers for their constructive feedback, and Rob Burbea for inspiration and support. 

\bibliography{library}

\appendix
\onecolumn
\aistatstitle{Do Bayesian Neural Networks Need To Be Fully Stochastic?\\Supplementary Materials}
\section{Proofs}\label{sec:supplement:proofs}
We provide a proof of Theorem~\ref{theorem:one_bias_is_all_you_need}, which states that a number of architectures are universal conditional distribution approximators (UCDAs). First, we restate the architectures that we consider and our theorem statement for convenience. The architectures that we consider are:
\begin{itemize}
\vspace{-5pt}
    \item[{[a]}] A deterministic multi-layer perceptron (MLP) with a single hidden layer of arbitrary width; non-polynomial activation function; and which takes $[Z;X]$ as its input.
    \item[{[b]}] An MLP with $L=2$ layers; continuous, invertible, and non-polynomial activation functions; $d$ units with deterministic biases and $m$ units with Gaussian random biases in the first layer; and a second layer of arbitrary width.
    \item[{[c]}] An MLP with $L=2$ layers; RELU activations; $2d$ units with deterministic biases and $m$ units with Gaussian random biases in the first layer; and a second layer of arbitrary width.
    \item[{[d]}] An MLP with $L\ge 2$ layers; continuous and non-polynomial activation functions that are either invertible or RELUs; at least $2\max(d+m,n)$ units with deterministic biases in each hidden layer; finite weights and biases throughout; one non-final hidden layer with $m$ additional units with Gaussian random biases (other layers may also have additional units with random biases, alongside their $2\max(d+m,n)$ deterministic ones), and; an arbitrary number of hidden units in one of the subsequent hidden layers.
    \vspace{-5pt}
\end{itemize}

We recall Theorem~\ref{theorem:one_bias_is_all_you_need}.
\ucda*

\vfill
\newpage

\begin{proof}
We start by noting that for any Gaussian $Z \in \mathbb{R}^m$, there must be some invertible matrix $V \in \mathbb{R}^{m\times m}$ and vector $u \in \mathbb{R}^m$ such that $Z = V \eta+u$, where $\eta \sim \mathcal{N}(0,I_m)$ can be used as the noise input to our generator function.
This is essentially a reparameterization, and it allows us to express $f_{\theta}(Z,x)$ as $f_{\theta}(V \eta + u,x)$.

We next show that if our network is able to represent the vector $[Z;x]$ exactly in one layer and the downstream subnetwork is a universal function approximator as per Lemma~\ref{theorem:uat}, this provides a sufficient condition for the result to hold.

More formally, assume that the all of the following hold for some hidden layer, $h_{\ell}\in \mathcal{H}_\ell \subset \mathbb{R}^\ell$,
\begin{enumerate}
    \item $Z$ and $x$ are fully input into the network by this layer;
    \item $h_{\ell}$ is compact provided $[Z;X]$ is itself is compact;
    \item $h_{\ell}$ can exactly represent $[Z;x]$ in the sense that there is some deterministic, surjective, and continuous function, $g : \mathcal{H}_\ell \rightarrow \mathbb{R}^m\times \mathcal{X}$, such that $g(h_{\ell})$ recovers $[Z;x]$ exactly for all $h_\ell$.
    \item The downstream network $f_{\theta}^{>\ell}(h_{\ell})$ satisfies the assumptions of Lemma~\ref{theorem:uat}.  
\end{enumerate}
Invoking Lemma~\ref{theorem:uat} for approximating the function $\tilde{f}\left([V^{-1};\mathbf{0}](g(h_{\ell})-[u;\mathbf{0}]),[\mathbf{0};I_d]g(h_{\ell})\right)=\tilde{f}(\eta,x)$ (noting that $\tilde{f}$ is continuous by assumption in the Theorem) gives
\begin{align}
  \forall \varepsilon>0,~ \exists \theta ~:~  \sup_{h_{\ell}\in\mathcal{H}_\ell} \|f_{\theta}^{>\ell}(h_{\ell}) - \tilde{f}\left([V^{-1};\mathbf{0}](g(h_{\ell})-[u;\mathbf{0}]),[\mathbf{0};I_d]g(h_{\ell})\right)\| < \varepsilon.
\end{align}
Now by the first assumption, $h_{\ell}$ must itself be a function of $[Z;x]=[V\eta+u;x]$, so we can rewrite the above as 
\begin{align*}
 \forall \varepsilon>0, \lambda<\infty~ \exists \theta ~:~  \sup_{x\in \mathcal{X},\eta\in \mathbb{R}^m, \|\eta\|<\lambda} \|f_{\theta}(V \eta +u,x) - \tilde{f}(\eta,x)\| < \varepsilon,
\end{align*}
which is the desired result, with $V$ and $u$ taking on the values required for $Z=V\eta+u$.
Here $\lambda$ and the assumption $\|\eta\|<\lambda$ have been introduced to ensure that $[Z;x]$ is itself compact, noting this further requires the assumption made in the theorem itself that $Z$ has finite mean and variance.

To complete the proof, we now need to show that the provided architectures are capable of producing networks that satisfy the four assumptions above.  

For architecture [a] they are all trivially satisfied as we have $h_0 = [Z;x]$, which directly ensures assumptions 1-3 hold, and $f_\theta^{>0}$ satisfies the assumptions of Lemma~\ref{theorem:uat} and is a suitable universal approximator.  

For architecture [b], we start by noting that the fourth assumption directly holds by the architecture construction.
Now by using the weight matrix $W_1=[\mathbf{0}; I_d]$ and the biases $b_1 = [Z;0]$ for this first layer, we have that its pre-activations are exactly $[Z;x]$ for all $Z$ and $x$.  
This ensures the first and second assumptions hold, noting that the continuity of the activation functions ensures that $h_{\ell}$ remains compact.
Finally, we can show that the third assumption holds by using the fact that the architecture uses invertible activation functions to simply define the required $g$ to be the corresponding inverse applied element-wise.

We can now view architecture [c] as an extension of architecture [b], wherein we no longer have an invertible activation function, but can exploit properties of the RELU and an increased number of hidden units instead.
Here we will now use the weight matrix $W_1=[\mathbf{0}; I_d; -I_d]$ and the biases $b_1 = [Z;\mathbf{0};\mathbf{0}]$ for this first layer, so that its pre-activations are exactly $[Z;x;-x]$ for all $Z$ and $x$.
This again immediately ensure that the first two assumption holds, while the fourth assumption is again immediately ensured by downstream subnetwork construction.
For the third assumption, we note that we have $h_{\ell}=[Z;\max(x,0);-\min(x,0)]$, and thus we already immediately have $Z$ and simply need to substract the third set of hidden units from the second to recover $x$, that is the assumptions is satisfied by taking $g([a;b;c])=[a;b-c]$.

Architecture [d] is now a generalization of those in [b] and [c] to allow additional layers and units in each layer.
We can show that the result holds for this set of architectures by showing that any such architecture can replicate the behavior of one of the architectures in [b] or [c] exactly.
For this, we first set all the weight matrices to the identity mapping and all the biases to zero for any layer which is not the specified layer with $m$ random Gaussian biases, with an arbitrary number of hidden units, or the output layer.
If the number of hidden units varies from one layer and the next, we simply pad the weight matrix with zeros, or truncate appropriately.
Here the assumption that we have at least $2\max(d+m,n)$ deterministic units in each layer means we always have enough units to exactly propagate either $[Z;x;-Z;-x]$ or $[Y;-Y]$, as required depending on the position in the network.
For the weights coming into the layer with the $m$ random biases, we use $W_{\ell}=[\mathbf{0}; I_d; -I_d; \mathbf{0}]$ and $b_{\ell} = [Z;\mathbf{0};\mathbf{0};\mathbf{0}]$, producing preactivations for $h_{\ell}$ that are always identical to the preactivations of $h_1$ in architecture [c], appended with zeros if necessary.
The arguments for architectures [b] and [c] (depending on whether our activations are invertible or RELUs) can now be applied to show that we can always recover $[Z;x]$ from $h_{\ell}$.
From here we simply note that the downstream network will behave identically as if it only had one more hidden layer of arbitrary width.
Thus, this architecture must always exactly emulate an architecture of type either [b] or [c], and is, therefore, a universal approximator as required.

\end{proof}

\section{Ethical Considerations}
\label{sec:supplement:ethics}
We hope that our work will help pave the way for cheap, high-quality uncertainty estimates. Such estimates could help build safe and robust artificial intelligence \cite{hendrycks2020unsolved}. Additionally, partially stochastic networks typically require less computation than fully stochastic networks and are therefore more environmentally friendly. However, strongly performing systems could lead to unintended consequences and pose societal costs \cite{russell2019human}, especially if humans place unwarranted credibility in the uncertainty estimates provided by deep learning systems. 

\section{Computational Considerations}
\label{sec:supplement:computational_considerations}
We now briefly discuss some of the computational considerations around partially stochasic networks.
At deployment, the memory cost of partially stochastic networks scales with the number of stochastic parameters; the fewer stochastic parameters used, the lower the memory cost, with the exact savings depending on the specific implementation. However, the cost of computing the subset predictive depends on the particular stochastic subset. For example, a stochastic input layer would \textit{not} reduce the number of forward passes required, whilst a stochastic output layer would. 

\vfill
\newpage

\newpage
\FloatBarrier
\section{Additional results and experiment details}

\FloatBarrier
\subsection{Fully Stochastic Networks with Bounded Variances are Not UCDAs}

In \S\ref{sec:theory}, we remarked that, at least in principle, fully stochastic networks can destroy required information about the inputs. We now demonstrate this empirically. 

We consider a 1D regression problem with synthetically generated data, and train a fully stochastic network and a partially stochastic network to match the predictive distribution of the dataset. Both networks use the same base architecture---a 2 hidden layer MLP with tanh activations---but the fully stochastic network maintains a distribution over all parameters with minimum standard deviation 0.25. In contrast, the partially stochastic network has one random bias in the input layer with fixed mean and variance. For training, we use moment matching: we optimise the output of the network to have the same mean and variance as the underlying data distribution. 

\begin{figure}[h]
    \centering
    \includegraphics[width=0.33\textwidth]{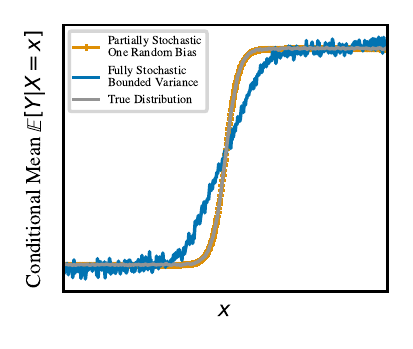}
    \caption{We train a fully stochastic MLP with bounded variance and a partially stochastic network with one random bias to match the mean and variance of a synthetic generated data distribution. Unlike the fully stochastic network, the partially stochastic network is able to match the conditional mean of underlying distribution.}
    \label{fig:expressivity}
\end{figure}

Fig.~\ref{fig:expressivity} shows the conditional mean for the fully stochastic network, the partially stochastic network, and the underlying data distribution. We see that the partially stochastic network is able to match the conditional mean of the underlying data distribution while the fully stochastic network is not.

\FloatBarrier
\subsection{HMC Mixing Analysis (\S\ref{sec:hmc_mixing})}
Here, we provide further results and details relating to the analysis in \S\ref{sec:hmc_mixing}: \nameref{sec:hmc_mixing} In this section, we analysed the convergence of HMC samples provided by \citet{izmailov2021bayesian}. Table~\ref{tab:hmc_details} contains details pertaining to this analysis.

\paragraph{Analysis Details}To compute the prediction associated with each chain, we averaged the softmax probabilities produced by the samples associated with the chain, in accordance with: 
\begin{align}
    p(y|x, \mathcal{D}) = \mathbb{E}_{p(\theta | \mathcal{D})}[p(y|x, \theta)].
\end{align}
That is, for each chain, we computed a predictive distribution by averaging the prediction probabilities for each class across the samples from the relevant chain. The ``prediction'' for each datapoint associated with each chain is the class that has the highest predictive probability for that i.e., $\arg \max_y p(y|x, \mathcal{D})$.

The agreement metric that we report is the percentage of data-points from a given dataset on which \textit{all three chains agree}. Note that this metric is different to the metric used by \citet{izmailov2021bayesian}, who compute the percentage of points on which one chain and the \textit{ensemble} of the remaining chains agree. 

\paragraph{Additional Results} Although we computed the agreement of each chain on all of the corruptions on the CIFAR-10-C dataset, we presented only a subset of corruptions in Fig.~\ref{fig:hmc_agreement}. Here, we additionally present results for the all corruptions below in Figure~\ref{fig:all_hmc_agreement}. 

\begin{figure}[h]
    \centering
    \includegraphics[width=\textwidth]{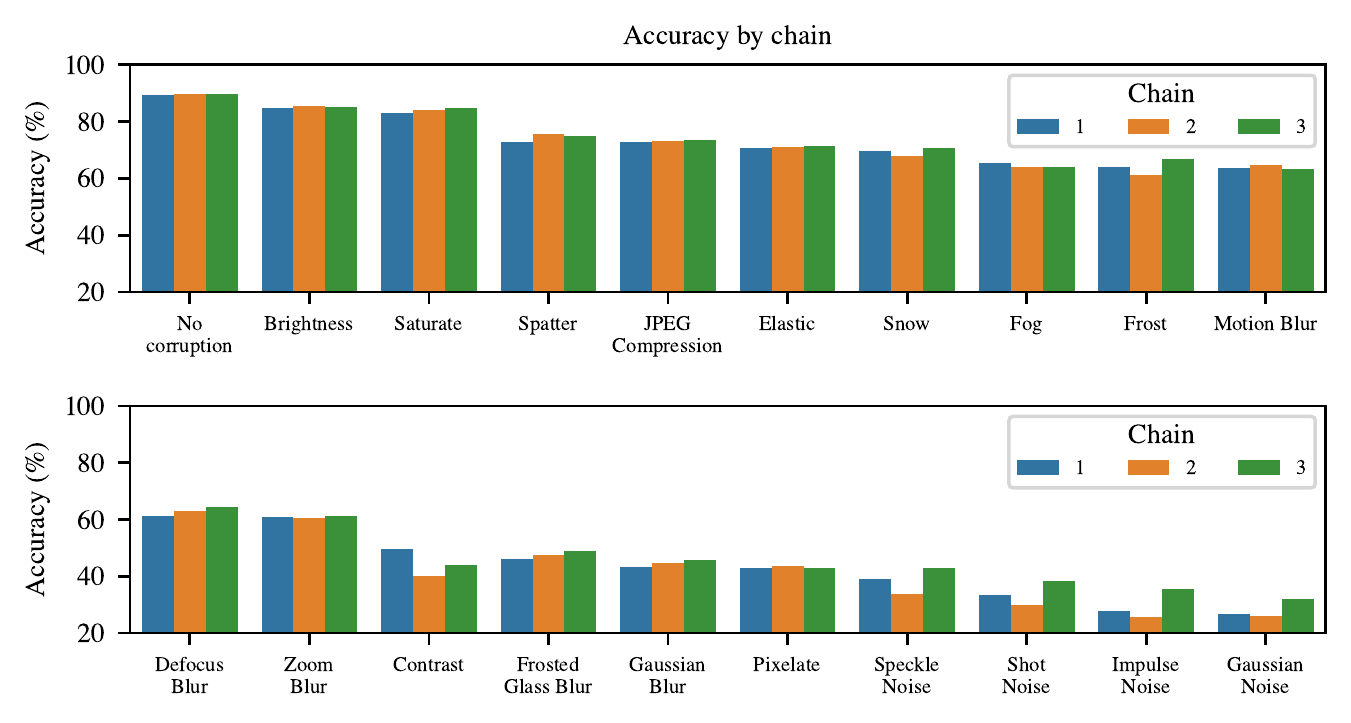}
    \caption{Assessment of function space mixing of ResNet-20-FRN full batch Hamiltonian Monte Carlo (HMC) samples trained on CIFAR-10. 
    We measure the variability in predictions made across HMC chains released by \citet{izmailov2021bayesian}. Here, we present the accuracy of each chain on the CIFAR-10 test set and all corruptions of the CIFAR-10-C \cite{hendrycks2018benchmarking} dataset with corruption intensity 5.}
    \label{fig:all_hmc_acc}
\end{figure}

\begin{figure}[h]
    \centering
    \includegraphics[width=\textwidth]{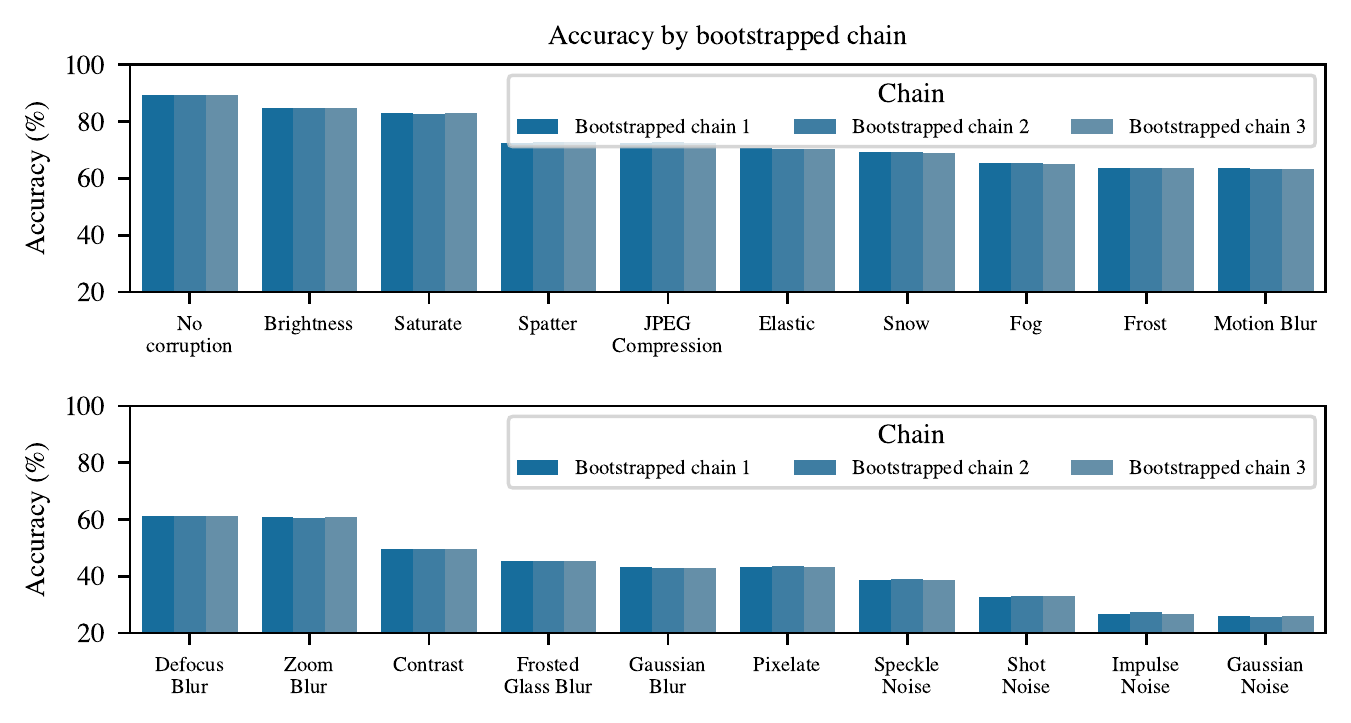}
    \caption{Assessment of within-chain function space variability of ResNet-20-FRN full batch Hamiltonian Monte Carlo (HMC) samples trained on CIFAR-10. We measure the variability in predictions made across simulated HMC chains, using released by \citet{izmailov2021bayesian}. Specifically, we generated multiple simulated chains by sampling from the first chain with replacement.}
    \label{fig:all_hmc_acc_bootstrapped}
\end{figure}

In an additional analysis, we compute the accuracy of each chain on different corruptions (Fig~\ref{fig:all_hmc_acc}). We find differences in accuracy of up to 8\% on certain corruptions, noticeably exceeding the within-chain variability (Fig~\ref{fig:all_hmc_acc_bootstrapped}). For example, the second HMC chain (orange) is less robust than the first and third HMC chain to all corruptions we consider. This further suggests that each HMC chain appears is exploring different regions of the posterior predictive. 


\begin{table}[h]
\centering
\caption{Additional details for analysis into whether full-batch HMC is converging, found in \S\ref{sec:hmc_mixing}: \nameref{sec:hmc_mixing}}
\vspace{5pt}
\begin{tabular}{@{}ll@{}}
\toprule
Hyper-parameter & Description \\ \midrule
Dataset & CIFAR-10 \citep{krizhevsky2009learning} (MIT license) \\
& CIFAR-10-C \citep{hendrycks2018benchmarking} (CC 4.0 license). \\
Use of existing assets & HMC samples from \cite{izmailov2021bayesian} (CC BY 4.0 license). \\
Architecture & ResNet-20-FRN, as in \cite{izmailov2021bayesian}. \\
Compute Infrastructure & Google Colab \\
Hardware & Tesla T4 (or Tesla P100). \\
Runtime & ca. 12 hours.\\ 
\midrule
\bottomrule
\end{tabular}
\label{tab:hmc_details}
\end{table}
 
\begin{figure}[h]
    \centering
    \includegraphics[width=\textwidth]{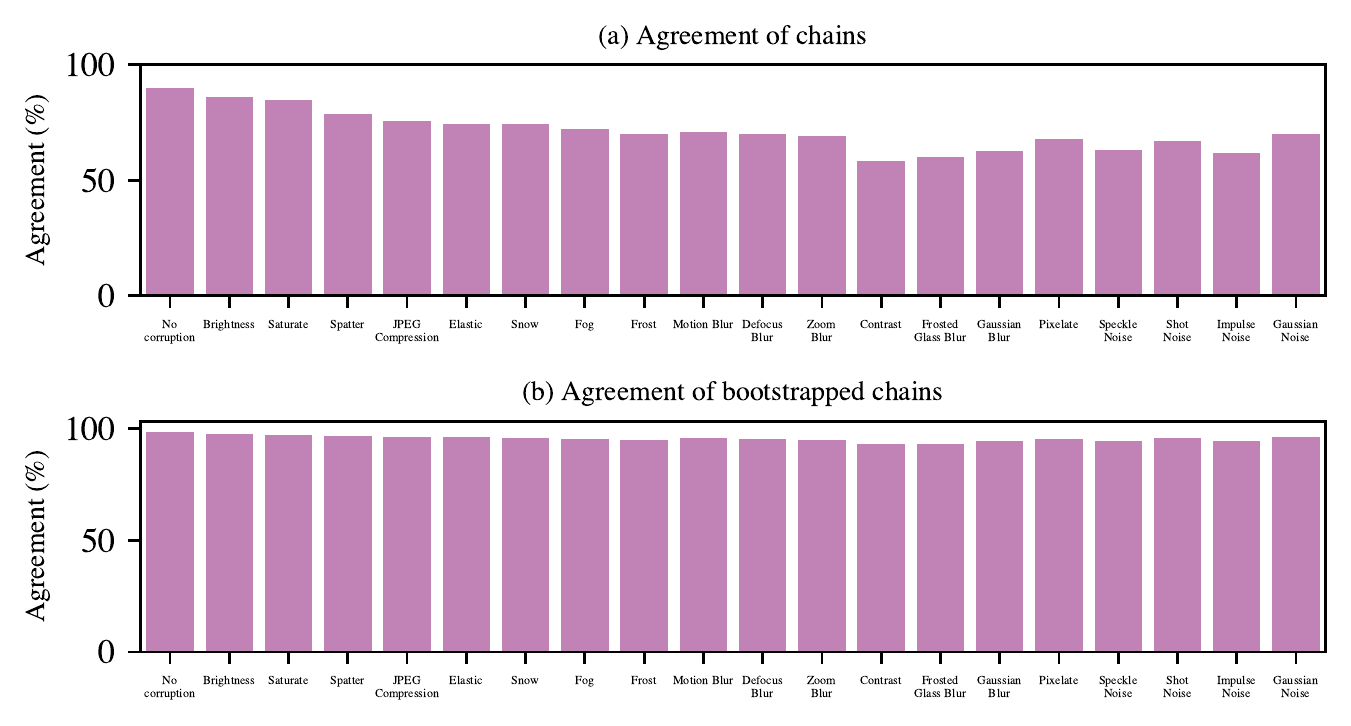}
    \caption{ Assessment of function space mixing of ResNet-20-FRN full batch Hamiltonian Monte Carlo (HMC) samples trained on CIFAR-10. 
    We measure the variability in predictions made across HMC chains released by \citet{izmailov2021bayesian}. To account for the finite sample size, we also measure the variability across simulated chains formed by resampling the first HMC chain i.e., bootstrapping. (a) We compute the percentage of points across different corruptions that all three chains make the same prediction on. While the agreement is 90\% on the CIFAR-10 test set, the agreement decreases to <60\% on certain datasets. (b) The agreement of bootstrapped HMC chains is greater than 94\% across all data considered.}
    \label{fig:all_hmc_agreement}
\end{figure} 

\FloatBarrier
\newpage
\subsection{1D Regression with Hamiltonian Monte Carlo (\S\ref{sec:1d_regression})}
We now provide further details relating to \S\ref{sec:1d_regression}: \nameref{sec:1d_regression}. In this section, we focus on the experiment details related to the experiments that used Hamiltonian Monte Carlo. Please see Table~\ref{tab:1d_regression_details_hmc} for relevant experiment details. 

\paragraph{Data} We generate synthetic data as follows. We draw 25 points from $\mathcal{U}(-3, -1.7)$ and 25 points from $\mathcal{U}(2.2, 4)$ to generate a set of 50 input points, $\{x_i\}$. We generate the output using $y_i = \sin(4\cdot(x_i-4.3)) + \epsilon_i$, where $\epsilon_i \sim \mathcal{N}(0, 0.05)^2$.

\begin{figure}
    \centering
    \vspace{-20pt}
    \includegraphics[width=\textwidth]{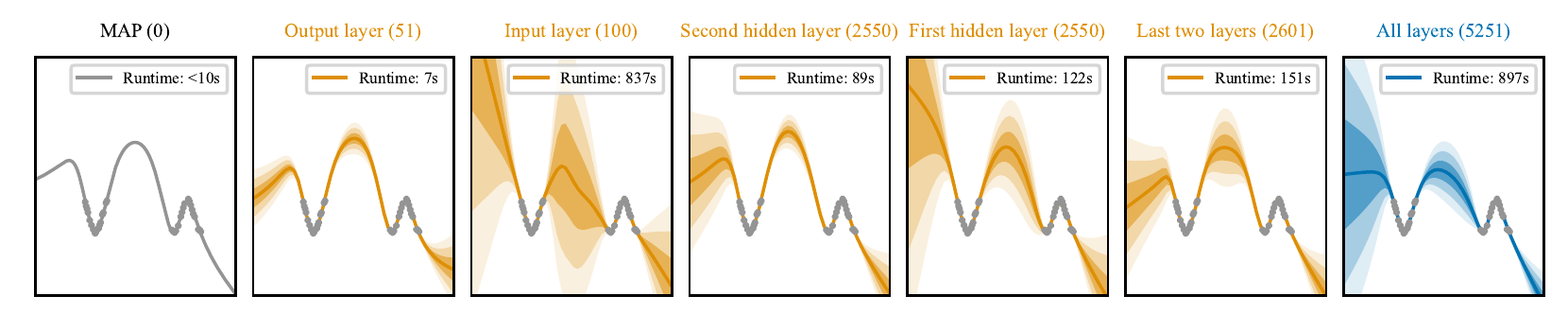}
    \caption{Additional partially stochastic network configurations using HMC inference over subsets of model parameters.}
    \vspace{-10pt}
    \label{fig:more_hmc_networks}
\end{figure}

\paragraph{Additional Results} In Fig.~\ref{fig:more_hmc_networks}, we show the predictive distributions of additional partially stochastic networks that use two-stage training. We note that for the the No-U-Turn Sampler (NUTS), the number of steps is chosen adaptively. 

\begin{table}[h]
\caption{Additional experiment details for 1D Regression using Hamiltonian Monte Carlo, found in \S\ref{sec:1d_regression}: \nameref{sec:1d_regression}.}
\vspace{5pt}
\centering
\begin{tabular}{@{}ll@{}}
\toprule
Hyper-parameter & Description \\ \midrule
Architecture & Multi-layer perceptron \\
Number of Hidden Layers & 2 \\ 
Layer Width & 50 \\
Activation Function & SiLU~\citep{hendrycks2016gaussian}\\
Prior Mean & 0 \\
Prior Variance & $\frac{|\Theta|}{|\Theta_S|}$, following \citep{daxberger2021bayesian}. \\
Network Parameterization & Neural Tangent Kernel Parameterization~\citep{jacot2018neural} \\
Inference Algorithm & Hamiltonian Monte Carlo~\citep{neal2012bayesian} with NUTS~\citep{no-u-turn} \\
MCMC chains & 8 \\
Warmup samples per chain & 1000 \\
Samples per chain & 500 \\
Maximum Tree Depth & 15 \\
Likelihood Function & Gaussian \\
Output Noise Variance & $0.05^2$ (As generated) \\
Dataset & Synthetic \\ 
Dataset Split & 70\% train, 20\% val, 10\% test. \\
Preprocessing & None \\ 
Computing Infrastructure & Macbook Pro \\
Runtime & ca. 15 minutes (Fully stochastic network). \\
\midrule
\bottomrule
\end{tabular}
\label{tab:1d_regression_details_hmc}
\end{table}

\FloatBarrier
\newpage
\subsection{1D Regression with Variational Inference (\S\ref{sec:1d_regression})}
We now provide further details relating to \S\ref{sec:1d_regression}: \nameref{sec:1d_regression}. In this section, we focus on the experiment details related to the experiments that used variational inference. Please see Table~\ref{tab:1d_regression_details} for relevant experiment details. 


\paragraph{Data} We generate synthetic data as follows. We draw 700 points from $\mathcal{U}(-2, -1.4)$ and 700 points from $\mathcal{U}(2, 2.8)$ to generate a set of 1400 input points, $\{x_i\}$. We generate the output using $y_i = \sin(4\cdot(x_i-4.3)) + \epsilon_i$, where $\epsilon_i \sim \mathcal{N}(0, 0.05)^2$.

\begin{table}[h]
\caption{Additional experiment details for 1d regression using variational inference, found in \S\ref{sec:1d_regression}: \nameref{sec:1d_regression}.}
\vspace{5pt}
\centering
\begin{tabular}{@{}ll@{}}
\toprule
Hyper-parameter & Description \\ \midrule
Architecture & Multi-layer perceptron \\
Number of Hidden Layers & 3 \\ 
Layer Width & 100 \\
Activation Function & Leaky ReLU \\
Prior & $\mathcal{N}(0, 1)$ \\
Training Monte Carlo Samples & 1 \\
Inference Algorithm & Flipout Mean-Field Variational Inference \citep{wen2018flipout} \\
Posterior Mean Initialisation & $\mu \sim \mathcal{N}(0, 0.1^2)$ \\
Posterior Standard Deviation Initialistion & $\sigma = \log(1+\exp(\rho))$, with $\rho \sim \mathcal{N}(-3, 0.1)$ \\
Stochastic Layers & All, or output layer only. \\
Likelihood Function & Gaussian \\
Output Noise Variance & $0.05^2$ (As generated) \\
Dataset & Synthetic \\ 
Dataset Split & 70\% train, 20\% val, 10\% test. \\
Preprocessing & None \\ 
Optimizer & AdamW \citep{loshchilov2017decoupled}\\
Learning Rate & $0.001$\\
Weight Decay & $0.0001$ only on deterministic weights and biases \\
Batch Zize & 350 \\
Epochs & $12000$\\
Plotting Epoch & Maximum validation set likelihood \\
Computing Infrastructure & Nvidia Tesla V100-PCIE-32GB \\
Runtime & ca. 15 minutes. \\
Use of existing assets & Bayesian Torch (BSD-3-Clause License)
\citep{krishnan2022bayesiantorch} \\
\midrule
\bottomrule
\end{tabular}
\label{tab:1d_regression_details}
\end{table}

\FloatBarrier
\newpage
\subsection{UCI Regression with Hamiltonian Monte Carlo (\S\ref{sec:uci_regression})}
We now provide further details relating to \S\ref{sec:uci_regression}: \nameref{sec:uci_regression}. Please see Table~\ref{tab:uci_regression_details} for relevant experiment details. 

\paragraph{Additional Details.} We note the additional details used in these experiments. (i) We used a homoscedastic noise model $p(y_i|x_i, \theta) = \mathcal{N}(y_i|f_\theta(x_i), \sigma_o^2)$, where $f_\theta(x_i)$ represents the neural network predictions. (ii) We tuned the prior variance so that the deterministic MAP network does not overfit. (iii) For the energy dataset, we predict only the first outcome variance, such that all the tasks we consider have one dimensional targets. (iv) All stochastic networks use a tempered posterior, where the sampler targets the density $\lambda\cdot\log p(\mathcal{D}|\theta) + \log p(\theta)$. We tuned $\lambda$ for each dataset by maximising the likelihood of a validation set. (v) We place a prior over the output noise precision, $\lambda_o = 1/\sigma_o^2$. 

\begin{table}[h]
\caption{Additional experiment details for UCI regression using Hamiltonian Monte Carlo, found in \S\ref{sec:uci_regression}: \nameref{sec:uci_regression}.}
\vspace{5pt}
\centering
\begin{tabular}{@{}ll@{}}
\toprule
Hyper-parameter & Description \\ \midrule
Architecture & Multi-layer perceptron \\
Number of Hidden Layers & 2 \\ 
Layer Width & 50 \\
Activation Function & Leaky ReLU \\
Prior & $\mathcal{N}(0, \sigma^2)$ \\
Prior Variance & $\sigma^2\in[0.1, 0.01, 0.01]$ for UCI Yacht, Boston and Energy respectively.\\
Likelihood Scale & $\lambda\in[6.0, 1.0, 8.0]$ for UCI Yacht, Boston and Energy respectively.\\
Inference Algorithm & Hamiltonian Monte Carlo~\citep{neal2012bayesian} with NUTS~\citep{no-u-turn}\\
MCMC chains & 8 \\
Warmup samples per chain & 325 \\
Samples per chain & 75 \\
Maximum Tree Depth & 15 \\
Output Precision Prior & $\operatorname{Gamma}(3.0, 1.0)$ \\ 
Likelihood Function & Gaussian \\
Datasets & UCI Yacht, Boston, Energy~\citep{Dua:2019} \\ 
Dataset Split & 90\% train, 10\% test. Standard and ``gap'' splits~\citep{foong2019between} \\
Preprocessing & Feature normalisation \\ 
Computing Infrastructure & Internal CPU Cluster \\
Runtime & $\leq$30 minutes; exact time depends on network. \\
\midrule
\bottomrule
\end{tabular}
\label{tab:uci_regression_details}
\end{table}

\FloatBarrier
\clearpage
\newpage
\subsection{Image Classification with Laplace Approximation (\S\ref{sec:vision_laplace})}
We now provide further results and details relating to \S\ref{sec:vision_laplace}: \nameref{sec:vision_laplace}. In this section, we considered the use of the Laplace approximation for fully stochastic and partially stochastic networks on an image classification task. Please see Table~\ref{tab:laplace_details} for relevant experiment details.

Note that the experiments in this section build heavily on the \texttt{Laplace} library, released by \citet{daxberger2021laplace}.

\begin{table}[h]
\centering
\caption{Additional experiment details for image classification experiments using the Laplace approximation, found in \S\ref{sec:vision_laplace}: \nameref{sec:vision_laplace}.}
\begin{tabular}{@{}ll@{}}
\toprule
Hyper-parameter & Description \\ \midrule
Architecture & FixUp \citep{zhang2019fixup} WideResNet-16-4 \citep{zagoruyko2016wide}\\ & following \citep{daxberger2021laplace}\\
Dataset & CIFAR-10 \citep{krizhevsky2009learning} (MIT License), \\
& CIFAR-10-C \citep{hendrycks2020unsolved} (CC 4.0 License). \\
Use of Existing Assets & \texttt{Laplace} Library \citep{daxberger2021laplace} (MIT License) \\
Computing Infrastructure & 4x Nvidia A100 GPU.\\
Preprocessing & Per-channel normalisation $\mu=0$, $\sigma=1$ \\ 
Number of Seeds & 10 \\
\midrule
\textbf{MAP Training} & \\ \midrule
Data Augmentation & Random crop and horizontal flip\\
Runtime & ca. 2 hours. \\
Epochs & 350 \\
Batch Size & 1024 \\
Optimizer & AdamW \citep{loshchilov2017decoupled}\\
Learning Rate & $0.001$\\
Weight Decay & $0.0001$ \\
\midrule
\textbf{Laplace Approximation} & \\ \midrule
Hessian Structure & Kronecker Factorised (KFAC) \\ 
Validation Set & $10\%$ of CIFAR-10 test set. \\
Prior Precision Tuning & Min val NLL (log-sweep in $(10^{-2}, 10^{5})$ with 125 increments)\\ 
Batch Size & 32 \\
Predictive & Linearized GLM Predictive \\ 
Temperature & 1.0 \\
Runtime & ca. 5 hours for fully stochastic networks \\
& less for partially stochastic networks \\ 
\midrule
\bottomrule
\end{tabular}
\label{tab:laplace_details}
\end{table}

\begin{figure}
    \centering
    \includegraphics[width=0.5\textwidth]{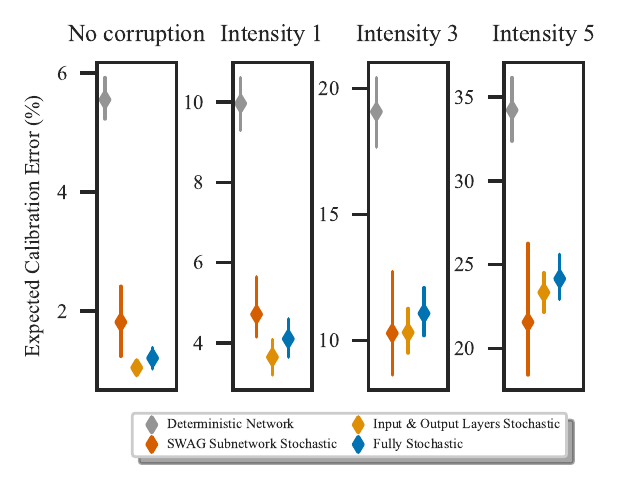}
    \caption{Calibration of Laplace approximation networks on CIFAR-10 and CIFAR-10-C. We compute the expected calibration error (ECE) for different Laplace approximation networks. Results are averaged across corruptions and shown for different corruption intensities. Markers and lines show mean and std. over 10 seeds.}
    \label{fig:laplace_calibration}
\end{figure}

\paragraph{Calibration of Laplace approximation networks} Our main results measure the relative performance of different Laplace approximation networks in terms of the negative log likelihood. Here, we additionally assess the quality of uncertainty estimates of different networks in terms of the expected calibration error (ECE). In Fig.~\ref{fig:laplace_calibration}, we see the calibration error increases as the input data is further corrupted, and further that partially stochastic networks can be better calibrated than fully stochastic ones.

\FloatBarrier
\clearpage
\newpage
\subsection{Image Classification with SWAG (\S\ref{sec:vision_swag})}
We now provide further results and details relating to \S\ref{sec:vision_swag}: \nameref{sec:vision_swag}. In this section, we considered the use of the SWAG inference for fully stochastic and partially stochastic networks on an image classification task. Please see Table~\ref{tab:swag_details} for relevant experiment details. We mostly followed \citet{maddox2019simple} in the choice of hyperparameters, using the hyperparameters they used for their ImageNet experiments from a pre-trained solution. We, however, tuned the learning rate per architecture using a validation set.

\begin{figure}
    \centering
    \includegraphics[width=\textwidth]{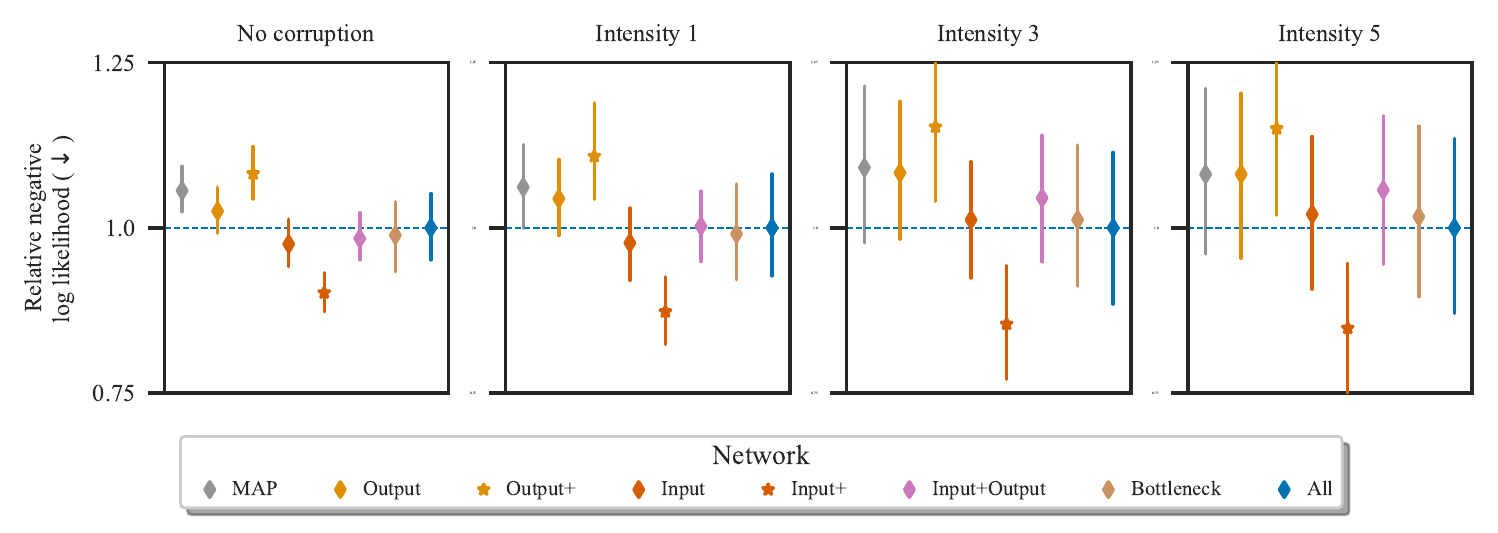}
    \caption{Relative NLL for various SWAG networks on CIFAR-10 and CIFAR-10-C \cite{hendrycks2018benchmarking}. Results averaged across 10 random seeds. We show many more configurations here---see Table~\ref{tab:swag_moreconfig_names} for correspondence between model name and the stochastic units.}
    \label{fig:swag_results_moreconfig}
\end{figure}

\paragraph{Additional Partially Stochastic Network Configurations} We present selected partially stochastic network configurations in Fig.~\ref{fig:swag_results}.  Fig.~\ref{fig:swag_results_moreconfig} shows more configurations. Several configurations outperform the fully stochastic network in distribution, but only the input and first ResNet block stochastic network outperforms the fully stochastic network on large corruption intensities. Nevertheless, the partially stochastic networks have lower memory cost.

\begin{table}[h]
\centering
\caption{Additional experiment details for image classification experiments using SWAG, found in \S\ref{sec:vision_swag}: \nameref{sec:vision_swag}}
\vspace{5pt}
\begin{tabular}{@{}ll@{}}
\toprule
Hyper-parameter & Description \\ \midrule
Architecture & FixUp \citep{zhang2019fixup} WideResNet-16-4 \citep{zagoruyko2016wide}\\ & following \citep{daxberger2021laplace}\\
Dataset & CIFAR-10 \citep{krizhevsky2009learning} (MIT License), \\
& CIFAR-10-C \citep{hendrycks2020unsolved} (CC 4.0 License). \\
Use of Existing Assets & \texttt{Laplace} Library \citep{daxberger2021laplace} (MIT License) \\
Computing Infrastructure & 4x Nvidia A100 GPU.\\
Preprocessing & Per-channel normalisation $\mu=0$, $\sigma=1$ \\ 
Number of Seeds & 10 \\
\midrule
\textbf{MAP Training} & \\ \midrule
Data Augmentation & Random crop and horizontal flip\\
Runtime & ca. 2 hours. \\
Epochs & 350 \\
Batch Size & 1024 \\
Optimizer & AdamW \citep{loshchilov2017decoupled}\\
Learning Rate & $0.001$\\
Weight Decay & $0.0001$ \\
\midrule
\textbf{SWAG} & \\ \midrule
Rank of Covariance Matrix ($K$) & 20 \\
Evaluation Monte Carlo Samples & 30 \\
SWAG Epochs & 10 \\
SWAG Snapshots per Epoch & 4 \\
Weight decay & 3e-4 \\
Validation Set & $10\%$ of CIFAR-10 test set. \\
Learning Rate & Tuned: log-sweep in $(10^{-5}, 10^{-2})$ with 25 increments)\\ 
Batch Size & 1024 \\
Runtime & ca. 3 hours \\ 
\midrule
\bottomrule
\end{tabular}
\label{tab:swag_details}
\end{table}

\begin{table}
\caption{Correspondence between network name and stochastic blocks for additional configurations for SWAG experiments (Fig.~\ref{fig:swag_results_moreconfig}). Note that ResNet block 1 is the ResNet block immediately after the input layer, and as the block number increases, the block is closer to the network output}
\centering
\begin{tabular}{@{}ll@{}}
\toprule
Name & Stochastic Units \\ \midrule
MAP & None \\
All (Fully Stochastic) & All layers \\
Input Layer & Input Layer \\
Input+ & Input Layer and ResNet Block 1\\
Output Layer & Output Layer \\
Output+ & Output Layer and ResNet Block 3\\ 
Input and Output Layer & Input and Output Layer \\
Bottleneck & ResNet Block 2 \\ 
\midrule
\bottomrule
\end{tabular}
\label{tab:swag_moreconfig_names}
\end{table}

\begin{figure}
    \centering
    \includegraphics[width=0.5\textwidth]{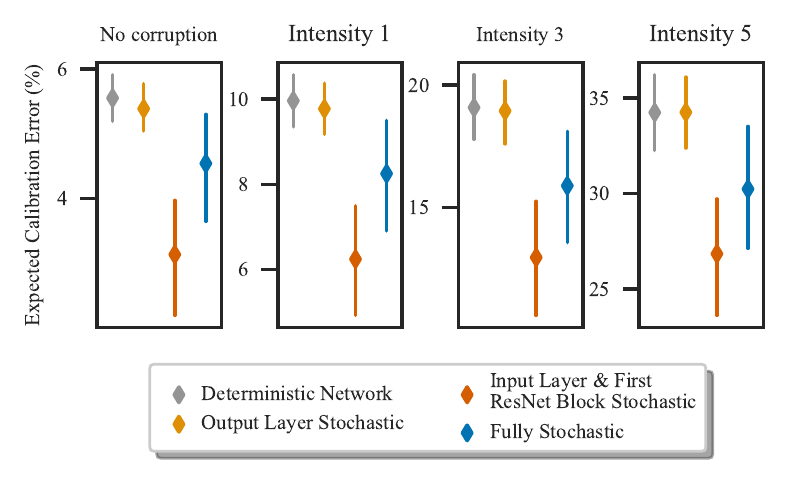}
    \caption{Calibration of SWAG inference networks on CIFAR-10 and CIFAR-10-C. We compute the expected calibration error (ECE) for different SWAG inference networks. Results are averaged across corruptions and shown for different corruption intensities. Markers and lines show mean and std. over 10 seeds.}
    \label{fig:swag_calibration}
\end{figure}

\paragraph{Calibration of SWAG inference networks} Our main results measure the relative performance of different SWAG inference networks in terms of the negative log likelihood. Here, we additionally assess the quality of uncertainty estimates of different networks in terms of the expected calibration error (ECE). In Fig.~\ref{fig:swag_calibration}, we see the calibration error increases as the input data is further corrupted, and further that partially stochastic networks can be better calibrated than fully stochastic ones.

\FloatBarrier
\clearpage
\newpage
\FloatBarrier
\subsection{Image Classification with Variational Inference}
We now provide further results and details relating to \S\ref{sec:vision_vi}: \nameref{sec:vision_vi}. In this section, we considered the use of variational inference for fully stochastic and partially stochastic networks on an image classification task. Please see Table~\ref{tab:vi_vision_details} for relevant experiment details.

Note that the experiments in this section build heavily on the \texttt{uncertainty-baselines} library, released by \citet{nado2021uncertainty}.

\begin{table}[h]
\centering
\caption{Additional experiment details for image classification experiments using variational inference,found in \S\ref{sec:vision_vi}:\nameref{sec:vision_vi}.}
\vspace{5pt}
\begin{tabular}{@{}ll@{}}
\toprule
Hyper-parameter & Description \\ \midrule
Architecture & WideResNet-28-10~\cite{zagoruyko2016wide}\\
Dataset & CIFAR-10, CIFAR-100~\cite{krizhevsky2009learning} (MIT License) \\
Use of Existing Assets & \texttt{uncertainty-baselines} \cite{nado2021uncertainty} (Apache 2.0 license) \\
Computing Infrastructure & 4x Nvidia A100 GPU. \\
Inference Algorithm & Flipout Mean-Field Variational Inference \cite{wen2018flipout}. \\
KL Annealing Epochs & 200 \\
Prior $\sigma$ & 0.1 \\
Posterior Standard Deviation Initialisation & 0.001\\
Training Monte Carlo Samples & 1 \\
Evaluation Monte Carlo Samples & 5 \\
Training Epochs & 250 \\
Dataset Split & 95\% train, 5\% validation. \\ 
$\ell_2$ Weight Decay & $4 \cdot 10^4$ \\
Batch Size & 256 \\ 
Learning Rate & 0.2 \\
Learning Rate Warmup Epochs & 1 \\
Momentum & 0.9 \\
Learning Rate Decay Ratio & 0.2 \\ 
Learning Rate Decay Epochs & 60, 120, 160 \\ 
Optimizer & SGD \\
Preprocessing & Per-channel normalisation $\mu=0$, $\sigma=1$ \\ 
Runtime & ca. 8 hours (fully stochastic) \\
\midrule
\bottomrule
\end{tabular}
\label{tab:vi_vision_details}
\end{table}

\begin{figure}
    \centering
    \includegraphics[width=\textwidth]{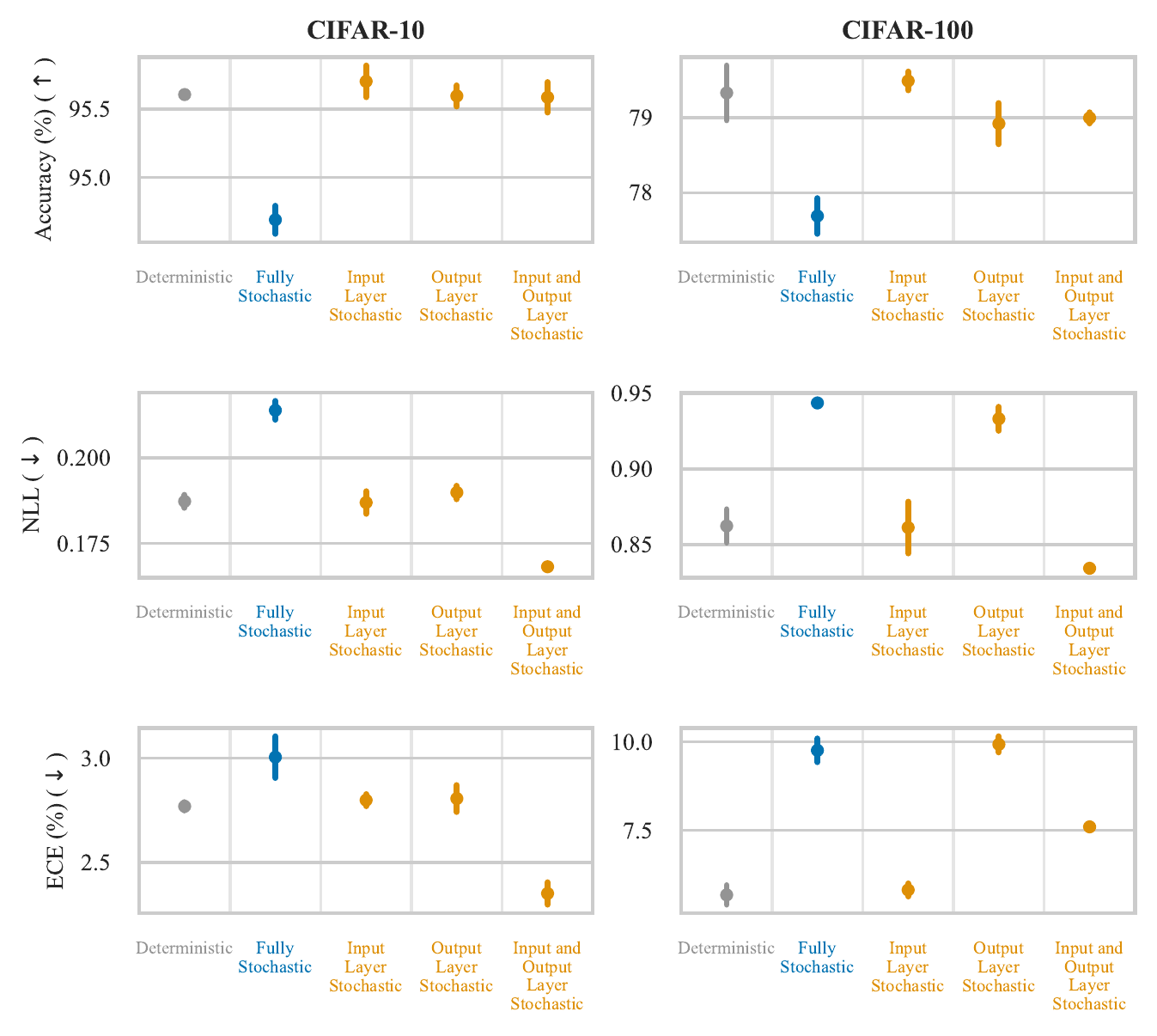}
    \caption{We report the accuracy, expected calibration error (ECE) and NLL on the standard CIFAR test sets when performing VI for subsets of parameters and learning the remaining parameters by maximising the (penalised) ELBO. Dots indicate the mean across 3 random seeds, bars indicate the standard deviation. This results are a graphical display of Table~\ref{tab:vi_results_table}, found in \S\ref{sec:vision_vi}: \nameref{sec:vision_vi}.}
    \label{fig:vi_results_variation}
\end{figure}

\paragraph{Variability across random seeds.} Fig.~\ref{fig:vi_results_variation} shows the mean and standard deviation of across different random seeds for large scale image classification with variational inference on the CIFAR test sets. The conclusions in \S\ref{sec:vision_vi}: \nameref{sec:vision_vi} are consistent across random seeds---partially stochastic networks can perform well, while fully stochastic networks do not appear to be well-performing despite their large computational cost.  

\begin{figure}
    \centering
    \includegraphics[width=\textwidth]{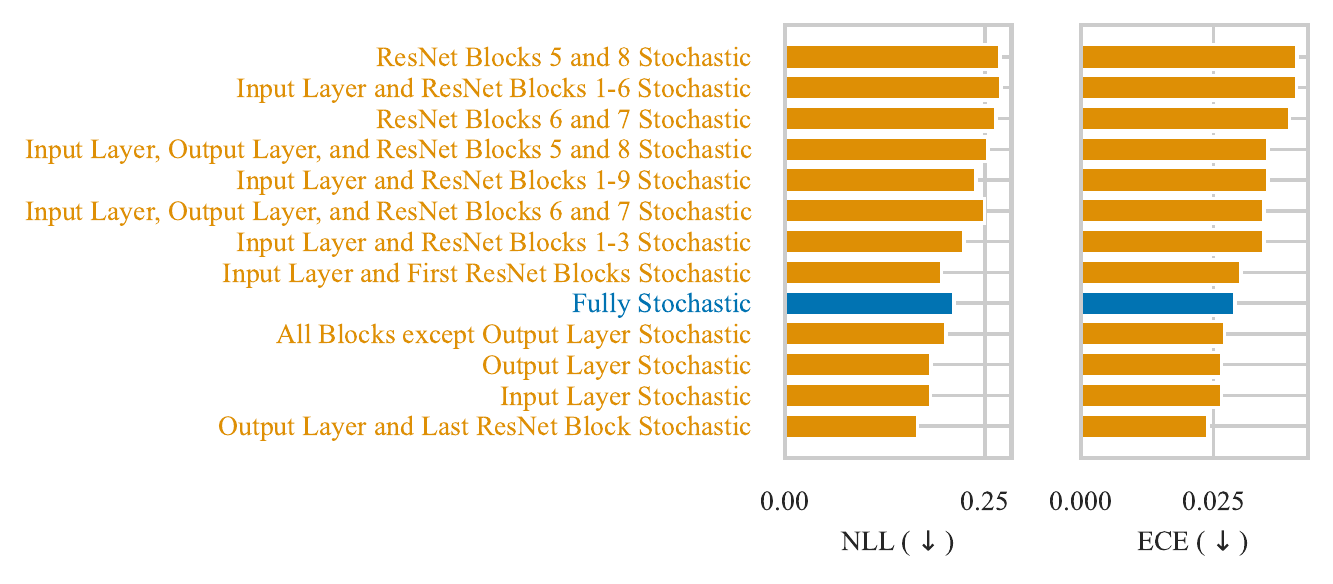}
    \caption{NLL and expected calibration error (ECE) on the CIFAR-10 test set for different network configurations. These results produced using only 1 random seed. Though every partially stochastic network does not perform well, there are performant partially stochastic networks.}
    \label{fig:vi_more_configs}
\end{figure}

\paragraph{Additional network configurations.} We considered several partially stochastic network considerations---see Fig.~\ref{fig:vi_more_configs}---and presented a selection of the results in \S\ref{sec:vision_vi}: \nameref{sec:vision_vi}. Though every partially stochastic network does not perform well, there are performant partially stochastic networks. One exciting area for future work is investigating and establishing best practices for the configuration and training of such partially stochastic networks. 

\end{document}